# Uniform Deviation Bounds for Unbounded Loss Functions like $k$-Means

**Olivier Bachem**                                   OLIVIER.BACHEM@INF.ETHZ.CH
**Mario Lucic**                                              LUCIC@INF.ETHZ.CH
**S. Hamed Hassani**                                        HAMED@INF.ETHZ.CH
**Andreas Krause**                                         KRAUSEA@ETHZ.CH
Department of Computer Science, ETH Zurich

## Abstract

Uniform deviation bounds limit the difference between a model's expected loss and its loss on an empirical sample *uniformly* for all models in a learning problem. As such, they are a critical component to empirical risk minimization. In this paper, we provide a novel framework to obtain uniform deviation bounds for loss functions which are *unbounded*. In our main application, this allows us to obtain bounds for $k$-Means clustering under weak assumptions on the underlying distribution. If the fourth moment is bounded, we prove a rate of $\mathcal{O}\left(m^{-\frac{1}{2}}\right)$ compared to the previously known $\mathcal{O}\left(m^{-\frac{1}{4}}\right)$ rate. Furthermore, we show that the rate also depends on the kurtosis — the normalized fourth moment which measures the "tailedness" of a distribution. We further provide improved rates under progressively stronger assumptions, namely, bounded higher moments, subgaussianity and bounded support.

## 1. Introduction

*Empirical risk minimization* — i.e. the training of models on a finite sample drawn i.i.d from an underlying distribution — is a central paradigm in machine learning. The hope is that models trained on the finite sample perform provably well even on previously unseen samples from the underlying distribution. But how many samples $m$ are required to guarantee a low approximation error $\epsilon$? *Uniform deviation bounds* provide the crucial answer. Informally, they are the worst-case difference across all possible models between the empirical loss of a model and its expected loss. As such, they determine how many samples are required to achieve a fixed error in terms of the loss function. In this paper, we consider the popular $k$-Means clustering problem and provide uniform deviation bounds based on *weak* assumptions on the underlying data generating distribution.

**Related work.** Traditional *Vapnik-Chervonenkis* theory provides tools to obtain uniform deviation bounds for binary concept classes such as classification using halfspaces (Vapnik & Chervonenkis, 1971). While these results have been extended to provide uniform deviation bounds for sets of continuous functions bounded in $[0, 1]$ (Haussler, 1992; Li et al., 2001), these results are not easily applied to $k$-Means clustering as the underlying loss function in $k$-Means clustering is continuous and unbounded.

In his seminal work, Pollard et al. (1981) shows that $k$-Means clustering is *strongly consistent*, i.e., that the optimal cluster centers on the empirical sample converge almost surely to the optimal centers of the distribution under a weak assumption. This has sparked a long line of research on *cluster stability* (Ben-David et al., 2006; Rakhlin & Caponnetto, 2007; Shamir & Tishby, 2007; 2008) which investigates the convergence of optimal parameters both asymptotically and for finite samples.

The vector quantization literature offers insights into the convergence of empirically optimal quantizers in terms of the quantization error — the $k$-Means loss function. A minimax rate of $\mathcal{O}\left(m^{-\frac{1}{2}}\right)$ is known if the underlying distribution has bounded support (Linder et al., 1994; Bartlett et al., 1998). A better rate of $\mathcal{O}\left(m^{-1}\right)$ may be achieved for finite support (Antos et al., 2005) or under both bounded support and regularity assumptions (Levrard et al., 2013).

Ben-David (2007) provides a uniform convergence result for center based clustering under a bounded support assumption. Telgarsky & Dasgupta (2013) prove uniform deviation bounds for $k$-Means clustering if the underlying distribution satisfies moment assumptions. In particular, in the case of a bounded fourth moment, they show a rate of $\mathcal{O}\left(m^{-\frac{1}{4}}\right)$.

**Our contributions.** We provide a novel framework to obtain uniform deviation bounds for unbounded loss functions. It relies on the tail behavior of the underlying distribution and is based on a generalization of the Vapnik-Chervonenkis dimension. In our main application to $k$-Means, it provides uniform deviation bounds with a rate of $\mathcal{O}\left(m^{-\frac{1}{2}}\right)$ for finite samples under weak assumptions. In contrast to prior work, our bounds are all scale-invariant and hold for any set of $k$ cluster centers (not *only* for a restricted solution set). We show that convergence depends on the kurtosis of the underlying distribution, which is the normalized fourth moment and measures the "tailedness" of a distribution. If bounded higher moments are available, we provide improved bounds that depend upon the normalized higher moments and we sharpen them even further under the stronger assumptions of subgaussianity and bounded support.

## 2. Problem statement for $k$-Means

We first focus on our main application, uniform deviation bounds for $k$-Means clustering, and defer the (more technical) framework for unbounded loss functions to Section 5. We consider a $d$-dimensional Euclidean space. For any $x \in \mathbb{R}^d$ and finite set $Q \subset \mathbb{R}^d$, we define

$$\mathrm{d}(x,Q)^2 = \min_{q \in Q} \|x - q\|_2^2.$$

Furthermore, in a slight abuse of notation, for $x, y \in \mathbb{R}^d$, we set $\mathrm{d}(x,y)^2 = \mathrm{d}(x,\{y\})^2 = \|x - y\|_2^2$.

**Statistical $k$-Means.** Let $P$ be any distribution on $\mathbb{R}^d$ with $\mu = \mathbb{E}_P[x]$ and $\sigma^2 = \mathbb{E}_P\left[\mathrm{d}(x,\mu)^2\right] \in (0, \infty)$. For any set $Q \subset \mathbb{R}^d$ of $k \in \mathbb{N}$ cluster centers, the *expected quantization error* is given by

$$\mathbb{E}_P\left[\mathrm{d}(x,Q)^2\right].$$

The goal of the *statistical $k$-Means problem* is to find a set of $k$ cluster centers such that the expected quantization error is minimized.

**Empirical $k$-Means.** Let $\mathcal{X}$ denote a finite set of points in $\mathbb{R}^d$. The goal of the *empirical $k$-Means problem* is to find a set $Q$ of $k$ cluster centers in $\mathbb{R}^d$ such that the *empirical quantization error* $\phi_\mathcal{X}(Q)$ is minimized, where

$$\phi_\mathcal{X}(Q) = \frac{1}{|\mathcal{X}|} \sum_{x \in \mathcal{X}} \mathrm{d}(x,Q)^2.$$

**Empirical risk minimization.** In practical machine learning, one often wishes to solve a statistical learning problem such as the variant of $k$-Means introduced above. Frequently, one assumes that there exists an unknown data distribution $P$ and that one can only observe independent samples from this distribution. The *empirical risk minimization* approach aims to minimize the error on unseen samples from $P$ — i.e., the expected error — by solving the learning problem on a finite sample. However, to do this, one needs to relate both the empirical learning problem and the statistical learning problem.

The goal of this paper is to derive bounds on the absolute difference between the expected quantization error based on $P$ and the empirical quantization error based on $m$ independent samples from $P$. More formally, we wish to bound the deviation

$$\left|\phi_{\mathcal{X}_m}(Q) - \mathbb{E}_P\left[\mathrm{d}(x,Q)^2\right]\right|,$$

*uniformly* for all $Q \in \mathbb{R}^{d \times k}$ where $\mathcal{X}_m$ is a set of $m$ independent samples from $P$. If this difference is sufficiently small for a given $m$, one may solve the empirical $k$-Means problem and obtain provable guarantees on the expected quantization error. Ideally, such a bound decreases with $m$ and approaches zero as $m \to \infty$. In that case, it also provides minimal sample sizes to achieve any given approximation error.

## 3. Uniform deviation bounds for $k$-Means

A simple approach to obtain uniform deviation bounds would be to try to bound the deviation by an *absolute* error $\epsilon$, i.e., to require that

$$\left|\phi_{\mathcal{X}_m}(Q) - \mathbb{E}_P\left[\mathrm{d}(x,Q)^2\right]\right| \leq \epsilon \tag{1}$$

uniformly for a set of possible solutions (Telgarsky & Dasgupta, 2013). In this paper, we provide uniform deviation bounds of a more general form: For any distribution $P$ and a sample of $m = f(\epsilon, \delta, k, d, P)$ points, we require that with probability at least $1 - \delta$

$$\left|\phi_{\mathcal{X}_m}(Q) - \mathbb{E}_P\left[\mathrm{d}(x,Q)^2\right]\right| \leq \frac{\epsilon}{2}\sigma^2 + \frac{\epsilon}{2}\mathbb{E}_P\left[\mathrm{d}(x,Q)^2\right] \tag{2}$$

*uniformly* for all $Q \in \mathbb{R}^{d \times k}$. The terms on the right-hand side may be interpreted as follows: The first term based on the variance $\sigma^2$ corresponds to a *scale-invariant, additive approximation error*. The second term is a *multiplicative approximation error* that allows the guarantee to hold even for solutions $Q$ with a large expected quantization error.

There are three key reasons why we choose (2) over (1): First, (1) is not scale-invariant and may thus not hold for classes of distributions that are equal up to scaling. Second, (1) may not hold for an unbounded solution space, e.g. $\mathbb{R}^{d \times k}$. Third, we can always rescale $P$ to unit variance and restrict ourselves to solutions $Q$ with $\mathbb{E}_P\left[\mathrm{d}(x,Q)^2\right] \leq \sigma^2$. Then, (2) implies (1) for a suitable transformation of $P$.

**Importance of scale-invariance.** If we scale all the points in a data set $\mathcal{X}$ and all possible sets of solutions $Q$ by some $\lambda > 0$, then the empirical quantization error is scaled by $\lambda^2$. Similarly, if we consider the random variable $\lambda x$ where $x \sim P$, then the expected quantization error is scaled by $\lambda^2$. At the same time, the $k$-Means problem remains the same: an optimal solution of the scaled problem is simply a scaled optimal solution of the original problem. Crucially, however, it is impossible to achieve the guarantee in (1) for distributions that are equal up to scaling: Suppose that (1) holds for some error tolerance $\epsilon$, and sample size $m$ with probability at least $1 - \delta$. Consider a distribution $P$ and a solution $Q \in \mathbb{R}^{d \times k}$ such that with probability at least $\delta$ we have

$$a < \left| \phi_{\mathcal{X}_m}(Q) - \mathbb{E}_P \left[ d(x, Q)^2 \right] \right|.$$

for some $a > 0$.[1] For $\lambda > \frac{1}{\sqrt{a\epsilon}}$, let $\tilde{P}$ be the distribution of the random variable $\lambda x$ where $x \sim P$ and let $\tilde{\mathcal{X}}_m$ consist of $m$ samples from $\tilde{P}$. Defining $\tilde{Q} = \{\lambda q \mid q \in Q\}$, we have with probability at least $\delta$

$$\left| \phi_{\tilde{\mathcal{X}}_m}\left(\tilde{Q}\right) - \mathbb{E}_{\tilde{P}} \left[ d\left(x, \tilde{Q}\right)^2 \right] \right| > a\lambda^2 > \epsilon$$

which contradicts (1) for the distribution $\tilde{P}$ and the solution $\tilde{Q}$. Hence, (1) cannot hold for both $P$ and its scaled transformation $\tilde{P}$.

**Unrestricted solution space** One way to guarantee *scale-invariance* would be require that

$$\left| \phi_{\mathcal{X}_m}(Q) - \mathbb{E}_P \left[ d(x, Q)^2 \right] \right| \leq \epsilon \sigma^2 \quad (3)$$

for all $Q \in \mathbb{R}^{d \times k}$. However, while (3) is scale-invariant it is also impossible to achieve for *all* solutions $Q$ as the following example shows. For simplicity, consider the 1-Means problem in 1 dimensional space and let $P$ be a distribution with zero mean. Let $\mathcal{X}_m$ denote $m$ independent samples from $P$ and denote by $\hat{\mu}$ the mean of $\mathcal{X}_m$. For any finite $m$, suppose that $\hat{\mu} \neq 0$ with high probability[2] and consider a solution $Q$ consisting of a single point $q \in \mathbb{R}$. We then have

$$\begin{aligned}
&\left| \phi_{\mathcal{X}_m}(\{q\}) - \mathbb{E}_P \left[ d(x, \{q\})^2 \right] \right| \\
&= \left| \phi_{\mathcal{X}_m}(\{\hat{\mu}\}) + d(\hat{\mu}, q)^2 - \sigma^2 - d(0, q)^2 \right| \quad (4) \\
&= \left| \phi_{\mathcal{X}_m}(\{\hat{\mu}\}) - \sigma^2 + q^2 - 2q\hat{\mu} + \hat{\mu}^2 - q^2 \right| \\
&= \left| \phi_{\mathcal{X}_m}(\{\hat{\mu}\}) - \sigma^2 + \hat{\mu}^2 - 2q\hat{\mu} \right|
\end{aligned}$$

Since $\hat{\mu} \neq 0$ with high probability, clearly this expression diverges as $q \to \infty$ and thus (3) cannot hold for arbitrary solutions $Q \in \mathbb{R}^{d \times k}$. Intuitively, the key issue is that both the empirical and the statistical error become unbounded as $q \to \infty$. Previous approaches such as Telgarsky & Dasgupta (2013) solve this issue by restricting the solution space from $\mathbb{R}^{d \times k}$ to solutions that are no worse than some threshold. In contrast, we allow the deviation between the empirical and the expected quantization error to scale with $\mathbb{E}_P \left[ d(x, Q)^2 \right]$.

**Arbitrary distributions.** Finally, we show that we either need to impose assumptions on $P$ or equivalently make the relationship between $m$, $\epsilon$ and $\delta$ in (2) depend on the underlying distribution $P$. Suppose that there exists a sample size $m \in \mathbb{N}$, an error tolerance $\epsilon \in (0, 1)$ and a maximal failure probability $\delta \in (0, 1)$ such that (2) holds for any distribution $P$. Let $P$ be the Bernoulli distribution on $\{0, 1\} \subset \mathbb{R}$ with $\mathbb{P}[x = 1] = p$ for $p \in (\delta^{\frac{1}{m}}, 1)$. By design, we have $\mu = p$, $\sigma^2 = p(1-p)$ and $\mathbb{E}_P \left[ d(x, 1)^2 \right] = (1-p)$. Furthermore, with probability at least $\delta$, the set $\mathcal{X}_m$ of $m$ independent samples from $P$ consists of $m$ copies of a point at one. Hence, (2) implies that with probability at least $1 - \delta$

$$\left| \phi_{\mathcal{X}_m}(1) - \mathbb{E}_P \left[ d(x, 1)^2 \right] \right| \leq \epsilon \mathbb{E}_P \left[ d(x, 1)^2 \right]$$

since $\sigma^2 \leq \mathbb{E}_P \left[ d(x, 1)^2 \right]$. However, with probability at least $\delta$, we have $\phi_{\mathcal{X}_m}(1) = 0$ which would imply $1 \leq \epsilon$ and thus lead to a contradiction with $\epsilon \in (0, 1)$.

## 4. Key results for $k$-Means

In this section, we present our main results for $k$-Means and defer the analysis and proofs to Sections 6.

### 4.1. Kurtosis bound

Similar to Telgarsky & Dasgupta (2013), the weakest assumption that we require is that the fourth moment of $d(x, \mu)$ for $x \in P$ is bounded.[3] Our results are based on the *kurtosis* of $P$ which we define as

$$\hat{M}_4 = \frac{\mathbb{E}_P \left[ d(x, \mu)^4 \right]}{\sigma^4}.$$

The kurtosis is the normalized fourth moment and is a scale-invariant measure of the "tailedness" of a distribution. For example, the normal distribution has a kurtosis of 2, while more heavy tailed distributions such as the $t$-Student distribution or the Pareto distribution have a potentially unbounded kurtosis. A natural interpretation of the kurtosis is provided by Moors (1986). For simplicity, consider a data set with unit variance. Then, the kurtosis may be restated as the shifted variance of $d(x, \mu)^2$, i.e.,

$$\hat{M}_4 = \text{Var}\left( d(x, \mu)^2 \right) + 1.$$

This provides a valuable insight into why the kurtosis is relevant for our setting: For simplicity, suppose we would like to estimate the expected quantization error $\mathbb{E}_P \left[ d(x, \mu)^2 \right]$ by the empirical quantization error $\phi_{\mathcal{X}_m}(\{\mu\})$ on a finite sample $\mathcal{X}_m$.[4] Then, the kurtosis measures the dispersion of $d(x, \mu)^2$ around its mean $\mathbb{E}_P \left[ d(x, \mu)^2 \right]$ and provides a bound on how many samples are required to achieve an error of $\epsilon$. While this simple example provides the key insight for the trivial solution $Q = \{\mu\}$, it requires a non-trivial effort to extend the guarantee in (2) to hold uniformly for all solutions $Q \in \mathbb{R}^{d \times k}$.

---

[1] For example, let $P$ be a nondegenerate multivariate normal distribution and $Q$ consist of $k$ copies of the origin.

[2] This holds for example if $P$ is the standard normal distribution.

[3] While our random variables $x \in P$ are potentially multivariate, it suffices to consider the behavior of the univariate random variable $d(x, \mu)$ for the assumptions in this section.

[4] This is a hypothetical exercise as $\mathbb{E}_P \left[ d(x, \mu)^2 \right] = 1$ by design. However, it provides an insight to the importance of the kurtosis.

With the use of a novel framework to learn unbounded loss functions (presented in Section 5), we are able to provide the following guarantee for $k$-Means.

**Theorem 1** (Kurtosis). *Let $\epsilon \in (0, 1)$, $\delta \in (0, 1)$ and $k \in \mathbb{N}$. Let $P$ be any distribution on $\mathbb{R}^d$ with Kurtosis $\hat{M}_4 < \infty$. For*

$$m \geq \frac{12800 \left(8 + \hat{M}_4\right)}{\epsilon^2 \delta} \left(3 + 30k(d+4) \log 6k + \log \frac{1}{\delta}\right)$$

*let $\mathcal{X} = \{x_1, x_2, \ldots, x_m\}$ be $m$ independent samples from $P$. Then, with probability at least $1 - \delta$, for all $Q \in \mathbb{R}^{d \times k}$*

$$\left|\phi_{\mathcal{X}}(Q) - \mathbb{E}_P\left[\mathrm{d}(x, Q)^2\right]\right| \leq \frac{\epsilon}{2}\sigma^2 + \frac{\epsilon}{2}\mathbb{E}_P\left[\mathrm{d}(x, Q)^2\right].$$

The proof is provided in Section 6.1. The number of required samples

$$m \in \Omega\left(\frac{\hat{M}_4}{\epsilon^2 \delta} \left(dk \log k + \log \frac{1}{\delta}\right)\right)$$

is linear in the kurtosis $\hat{M}_4$ and the dimensionality $d$, near-linear in the number of clusters $k$ and $\frac{1}{\delta}$, and quadratic in $\frac{1}{\epsilon}$. Intuitively, the bound may be interpreted as follows: $\Omega\left(\frac{\hat{M}_4}{\epsilon^2 \delta}\right)$ samples are required such that the guarantee holds for a single solution $Q \in \mathbb{R}^{d \times k}$. Informally, a generalization of the Vapnik Chervonenkis dimension for $k$-Means clustering may be bounded by $\mathcal{O}(dk \log k)$ and measures the "complexity" of the learning problem. The multiplicative $dk \log k + \log \frac{1}{\delta}$ term intuitively extends the guarantee *uniformly* to all possible $Q \in \mathbb{R}^{d \times k}$. We refer to Section 6.1 for a formal derivation of the bound.

We compare our results to the one obtained in Telgarsky & Dasgupta (2013) based on a fourth moment bound. While we require a bound on the normalized fourth moment, i.e. the kurtosis, Telgarsky & Dasgupta (2013) consider the case where all unnormalized moments up to the fourth are uniformly bounded by some $M$, i.e.,

$$\mathbb{E}_P\left[\mathrm{d}(x, \mu)^l\right] \leq M, \quad 1 \leq l \leq 4.$$

They provide uniform deviation bounds for all solutions $Q$ such that either $\phi_{\mathcal{X}}(Q) \leq c$ or $\mathbb{E}_P\left[\mathrm{d}(x, Q)^2\right] \leq c$ for some $c > 0$. To compare our bounds, we consider a data set with unit variance to compare the different bounds and restrict ourselves to solutions $Q \in \mathbb{R}^{d \times k}$ with an expected quantization error of at most the variance, i.e., $\mathbb{E}_P\left[\mathrm{d}(x, Q)^2\right] \leq \sigma^2 = 1$. We consider bounds on the maximal deviation

$$\Delta = \sup_{Q \in \mathbb{R}^{d \times k} : \mathbb{E}_P[\mathrm{d}(x, Q)^2] \leq 1} \left|\phi_{\mathcal{X}}(Q) - \mathbb{E}_P\left[\mathrm{d}(x, Q)^2\right]\right|.$$

Telgarsky & Dasgupta (2013) bound this deviation by

$$\Delta \in \mathcal{O}\left(\sqrt{\frac{M^2}{\sqrt{m}}\left(dk \log(Mdm) + \log \frac{1}{\delta}\right)} + \sqrt{\frac{1}{m\delta^2}}\right).$$

In contrast, our bound in Theorem 1 implies

$$\Delta \in \mathcal{O}\left(\sqrt{\frac{\hat{M}_4}{m\delta}\left(dk \log k + \log \frac{1}{\delta}\right)}\right).$$

The key difference is in how $\Delta$ scales with the sample size $m$. While Telgarsky & Dasgupta (2013) show a rate of $\Delta \in \mathcal{O}\left(m^{-\frac{1}{4}}\right)$, we improve it to $\Delta \in \mathcal{O}\left(m^{-\frac{1}{2}}\right)$.

### 4.2. Bounded higher moments

The tail behavior of $\mathrm{d}(x, \mu)$ may be characterized by the moments of $P$. Hence, if the underlying distribution has higher moments that are bounded, we are able to sharpen our bound. For $p \in \mathbb{N}$, we consider the *standardized $p$-th moment* of $P$, i.e.,

$$\hat{M}_p = \frac{\mathbb{E}_P\left[\mathrm{d}(x, \mu)^p\right]}{\sigma^p}.$$

**Theorem 2** (Moment bound). *Let $\epsilon \in (0, 1)$, $\delta \in (0, 1)$ and $k \in \mathbb{N}$. Let $P$ be any distribution on $\mathbb{R}^d$ with finite $p$-th order moment bound $\hat{M}_p < \infty$ for $p \in \{4, 8, \ldots, \infty\}$. For $m \geq \max\left(\frac{3200 m_1}{\epsilon^2}, \left(\frac{8}{\delta}\right)^{\frac{8}{p}}\right)$ with*

$$m_1 = p\left(4 + \hat{M}_p^{\frac{4}{p}}\right)\left(3 + 30k(d+4) \log 6k + \log \frac{1}{\delta}\right)$$

*let $\mathcal{X} = \{x_1, x_2, \ldots, x_m\}$ be $m$ independent samples from $P$. Then, with probability at least $1 - \delta$, for all $Q \in \mathbb{R}^{d \times k}$*

$$\left|\phi_{\mathcal{X}}(Q) - \mathbb{E}_P\left[\mathrm{d}(x, Q)^2\right]\right| \leq \frac{\epsilon}{2}\sigma^2 + \frac{\epsilon}{2}\mathbb{E}_P\left[\mathrm{d}(x, Q)^2\right].$$

The proof is provided in Section 6.2. Compared to the previous bound based on the kurtosis, Theorem 2 requires

$$m \in \Omega\left(\frac{p\hat{M}_p^{\frac{4}{p}}}{\epsilon^2}\left(dk \log k + \log \frac{1}{\delta}\right) + \left(\frac{1}{\delta}\right)^{\frac{8}{p}}\right)$$

samples. In particular, with higher order moment bounds, it is easier to achieve high probability results since the dependence on $\frac{1}{\delta}$ is only of $\Omega\left(\left(\frac{1}{\delta}\right)^{\frac{8}{p}}\right)$ compared to near linear for a kurtosis bound. The quantity $\hat{M}_p^{\frac{4}{p}}$ may be interpreted as a bound on the kurtosis $\hat{M}_4$ based on the higher order moment $\hat{M}_p$. In fact, Hoelder's inequality implies that $\hat{M}_4 \leq \hat{M}_p^{\frac{4}{p}}$. While the result only holds for $p \in \{8, 12, 16, \ldots, \infty\}$, it is trivially extended to $p' \geq 8$: Apply Theorem 2 with $p = 4 \left\lfloor \frac{p'}{4} \right\rfloor$ and note that by Hoelder's inequality $\hat{M}_p^{\frac{4}{p}} \leq \hat{M}_{p'}^{\frac{4}{p'}}$.

As in the previous subsection, we compare our results to Telgarsky & Dasgupta (2013) for distributions $P$ that have unit variance and we restrict ourselves to solutions $Q \in \mathbb{R}^{d \times k}$ with an expected quantization error of at most the variance, i.e., $\mathbb{E}_P\left[\mathrm{d}(x, Q)^2\right] \leq \sigma^2 = 1$. Telgarsky & Dasgupta (2013) require that there exists a bound $M$

$$\mathbb{E}_P\left[\mathrm{d}(x, \mu)^l\right] \leq M, \quad 1 \leq l \leq p.$$

Then for $m$ sufficiently large, the maximal deviation $\Delta$ is of

$$\mathcal{O}\left(\sqrt{\frac{M^{\frac{8}{p}}}{m^{1-\frac{4}{p}}}\left(dk \ln(M^{\frac{4}{p}} dm) + \ln \frac{1}{\delta}\right)} + \frac{2^{\frac{p}{4}}}{m^{\frac{3}{4}-\frac{2}{p}}}\left(\frac{1}{\delta}\right)^{\frac{4}{p}}\right).$$

In contrast, we obtain, for $m$ sufficiently large,

$$\Delta \in \mathcal{O}\left(\sqrt{\frac{p \hat{M}_p^{\frac{4}{p}}}{m}\left(dk \log k + \log \frac{1}{\delta}\right)}\right).$$

While Telgarsky & Dasgupta (2013) only show a rate of $\mathcal{O}\left(m^{-\frac{1}{2}}\right)$ as $p \to \infty$, we obtain a $\in \mathcal{O}\left(m^{-\frac{1}{2}}\right)$ rate for all higher moment bounds.

### 4.3. Subgaussianity

If the distribution $P$ is subgaussian, then all its moments $\hat{M}_p$ are bounded. By optimizing $p$ in Theorem 2, we are able to show the following bound.

**Theorem 3** (Subgaussian bound). *Let $\epsilon \in (0, 1)$, $\delta \in (0, 1)$ and $k \in \mathbb{N}$. Let $P$ be any distribution on $\mathbb{R}^d$ with $\mu = \mathbb{E}_P[x]$ and*

$$\forall t > 0: \quad \mathbb{P}\left[\mathrm{d}(x, \mu) > t\sigma\right] \leq a \exp\left(-\frac{t^2}{\sqrt{b}}\right)$$

*for some $a > 1, b > 0$. Let $m \geq \frac{3200 m_1}{\epsilon^2}$ with*

$$m_1 = p\left(4 + \frac{abp^2}{4}\right)\left(3 + 30k(d+4) \log 6k + \log \frac{1}{\delta}\right).$$

*and $p \leq 9 + 4 \log \frac{1}{\delta}$. Let $\mathcal{X} = \{x_1, x_2, \ldots, x_m\}$ be $m$ independent samples from $P$. Then, with probability at least $1 - \delta$, for all $Q \in \mathbb{R}^{d \times k}$*

$$\left|\phi_\mathcal{X}(Q) - \mathbb{E}_P\left[\mathrm{d}(x, Q)^2\right]\right| \leq \frac{\epsilon}{2}\sigma^2 + \frac{\epsilon}{2}\mathbb{E}_P\left[\mathrm{d}(x, Q)^2\right].$$

The proof is provided in Section 6.3. In $\Omega(\cdot)$ notation, we hence require

$$m \in \Omega\left(\frac{ab \log^3 \frac{1}{\delta}}{\sigma \epsilon^2}\left(dk \log k + \log \frac{1}{\delta}\right)\right)$$

samples. This result features a polylogarithmic dependence on $\frac{1}{\delta}$ compared to the polynomial dependence for the bounds based on bounded higher moments. The required sample size further scales linearly with the (scale-invariant) subgaussianity parameters $a$ and $b$. For example, if $P$ is a one-dimensional normal distribution of any scale, we would have $a = 2$ and $b = 1$.

### 4.4. Bounded support

Finally, the strongest assumption that we consider is if the support of $P$ is bounded by a hypersphere in $\mathbb{R}^d$ with diameter $R > 0$. This ensures that almost surely $\mathrm{d}(x, \mu) \leq R$ and hence $\hat{M}_4 \leq \frac{R^4}{\sigma^4}$. This allows us to obtain the following result.

**Theorem 4** (Bounded support). *Let $\epsilon \in (0, 1)$, $\delta \in (0, 1)$ and $k \in \mathbb{N}$. Let $P$ be any distribution on $\mathbb{R}^d$, with $\mu = \mathbb{E}_P[x]$ and $\sigma^2 = \mathbb{E}_P\left[\mathrm{d}(x, \mu)^2\right] \in (0, \infty)$, whose support is contained in a $d$-dimensional hypersphere of diameter $R > 0$. For*

$$m \geq \frac{12800\left(8 + \frac{R^4}{\sigma^4}\right)}{\epsilon^2}\left(3 + 30k(d+4) \log 6k + \log \frac{1}{\delta}\right)$$

*let $\mathcal{X} = \{x_1, x_2, \ldots, x_m\}$ be $m$ independent samples from $P$. Then, with probability at least $1 - \delta$, for all $Q \in \mathbb{R}^{d \times k}$*

$$\left|\phi_\mathcal{X}(Q) - \mathbb{E}_P\left[\mathrm{d}(x, Q)^2\right]\right| \leq \frac{\epsilon}{2}\sigma^2 + \frac{\epsilon}{2}\mathbb{E}_P\left[\mathrm{d}(x, Q)^2\right].$$

The proof is provided in Section 6.4. Again, the required sample size scales linearly with the kurtosis bound $\frac{R^4}{\sigma^4}$. However, the bound is only logarithmic in $\frac{1}{\delta}$.

## 5. Framework for unbounded loss functions

To obtain the results presented in Section 4, we propose a novel framework to uniformly approximate the expected values of a set of unbounded functions based on an empirical sample. We consider a function family $\mathcal{F}$ mapping from an arbitrary input space $\mathcal{X}$ to $\mathbb{R}_{\geq 0}$ and a distribution $P$ on $\mathcal{X}$. We further require a generalization of the *Vapnik-Chervonenkis dimension* to continuous, unbounded functions[5] — the *pseudo-dimension*.

---

[5]The pseudo-dimension was originally defined for sets of functions mapping to $[0, 1]$ (Haussler, 1992; Li et al., 2001). However, it is trivially extended to unbounded functions mapping to $\mathbb{R}_{\geq 0}$.

**Definition 1** (Haussler (1992); Li et al. (2001)). *The pseudo-dimension of a set $\mathcal{F}$ of functions from $\mathcal{X}$ to $\mathbb{R}_{\geq 0}$, denoted by $\mathrm{Pdim}(\mathcal{F})$, is the largest $d'$ such there is a sequence $x_1, \ldots, x_{d'}$ of domain elements from $X$ and a sequence $r_1, \ldots, r_{d'}$ of reals such that for each $b_1, \ldots, b_{d'} \in \{above, below\}$, there is an $f \in \mathcal{F}$ such that for all $i = 1, \ldots, d'$, we have $f(x_i) \geq r_i \iff b_i = above$.*

Similar to the VC dimension, the pseudo-dimension measures the cardinality of the largest subset of $\mathcal{X}$ that can be *shattered* by the function family $\mathcal{F}$. Informally, the pseudo-dimension measures the richness of $\mathcal{F}$ and plays a critical role in providing a uniform approximation guarantee across all $f \in \mathcal{F}$. With this notion, we are able to state the main result in our framework.

**Theorem 5.** *Let $\epsilon \in (0, 1)$, $\delta \in (0, 1)$ and $t > 0$. Let $\mathcal{F}$ be a family of functions from $\mathcal{X}$ to $\mathbb{R}_{\geq 0}$ with $\mathrm{Pdim}(F) = d < \infty$. Let $s : \mathcal{X} \to \mathbb{R}_{\geq 0}$ be a function such that $s(x) \geq \sup_{f \in \mathcal{F}} f(x)$ for all $x \in \mathcal{X}$. Let $P$ be any distribution on $\mathcal{X}$ and for*

$$m \geq \frac{200t}{\epsilon^2}\left(3 + 5d + \log\frac{1}{\delta}\right),$$

*let $x_1, x_2, \ldots, x_{2m}$ be $2m$ independent samples from $P$. Then, if*

$$\mathbb{E}_P\left[s(x)^2\right] \leq t \quad \text{and} \quad \mathbb{P}\left[\frac{1}{2m}\sum_{i=1}^{2m} s(x_i)^2 > t\right] \leq \frac{\delta}{4}, \tag{5}$$

*it holds with probability at least $1 - \delta$ that*

$$\left|\frac{1}{m}\sum_{i=1}^{m} f(x_i) - \mathbb{E}_P[f(x)]\right| \leq \epsilon, \quad \forall f \in \mathcal{F}. \tag{6}$$

Applying Theorem 5 to a function family $\mathcal{F}$ requires three steps: First, one needs to bound the pseudo-dimension of $\mathcal{F}$. Second, it is necessary to find a function $s : \mathcal{X} \to \mathbb{R}_{\geq 0}$ such that

$$f(x) \leq s(x), \quad \forall x \in \mathcal{X} \text{ and } \forall f \in \mathcal{F}.$$

Ideally, such a bound should be as tight as possible. Third, one needs to find some $t > 0$ and a sample size

$$m \geq \frac{200t}{\epsilon^2}\left(3 + 5d + \log\frac{1}{\delta}\right)$$

such that

$$\mathbb{E}_P\left[s(x)^2\right] \leq t \quad \text{and} \quad \mathbb{P}\left[\frac{1}{2m}\sum_{i=1}^{2m} s(x_i)^2 > t\right] \leq \frac{\delta}{4}.$$

Finding such a bound usually entails examining the tail behavior of $s(x)^2$ under $P$. Furthermore, it is evident that a bound $t$ may only be found if $\mathbb{E}_P\left[s(x)^2\right]$ is bounded and that assumptions on the distribution $P$ are required. In Section 6, we will see that for $k$-Means a function $s(x)$ with $\mathbb{E}_P\left[s(x)^2\right] < \infty$ may be found if the kurtosis of $P$ is bounded.

We defer the proof of Theorem 5 to Section B of the Supplementary Materials and provide a short proof sketch that captures the main insight.

*Proof sketch.* Our proof is based on a *double sampling* approach. Let $x_{m+1}, x_{m+2}, \ldots, x_{2m}$ be an additional $m$ independent samples from $P$ and let $\sigma_1, \sigma_2, \ldots, \sigma_m$ be independent random variables uniformly sampled from $\{-1, 1\}$. Then, we show that, if $\mathbb{E}_P\left[s(x)^2\right] \leq t$, the probability of (6) not holding may be bounded by the probability that there exists a $f \in \mathcal{F}$ such that

$$\left|\frac{1}{m}\sum_{i=1}^{m} \sigma_i \left(f(x_i) - f(x_{i+m})\right)\right| > \epsilon. \tag{7}$$

We first provide the intuition for a single function $f \in \mathcal{F}$ and then show how we extend it to all $f \in \mathcal{F}$. While the function $f(x)$ is not bounded, for a given sample $x_1, x_2, \ldots, x_{2m}$, each $f(x_i)$ is contained within $[0, s(x_i)]$. Given the sample $x_1, x_2, \ldots, x_{2m}$, the random variable $\sigma_i\left(f(x_i) - f(x_{i+m})\right)$ is bounded in $0 \pm \max\left(s(x_i), s(x_{i+m})\right)$ and has zero mean. Hence, given independent samples $x_1, x_2, \ldots, x_{2m}$, the probability of (7) occurring for a single $f \in \mathcal{F}$ can be bounded using Hoeffding's inequality by

$$2\exp\left(-\frac{2m\epsilon^2}{\frac{1}{m}\sum_{i=1}^{m}\max\left(s(x_i), s(x_{i+m})\right)^2}\right)$$

$$\leq 4\exp\left(-\frac{2m\epsilon^2}{\frac{1}{2m}\sum_{i=1}^{2m} s(x_i)^2}\right).$$

By (5), with probability at least $1 - \frac{\delta}{4}$, we have $\frac{1}{2m}\sum_{i=1}^{2m} s(x_i)^2 \leq t$ and we hence require $m \in \Omega\left(\frac{t\log\frac{1}{\delta}}{\epsilon^2}\right)$ samples to guarantee that (7) does not hold for a single $f \in \mathcal{F}$ with probability at least $1 - \frac{\delta}{4}$.

To bound the probability that there exists *any* $f \in \mathcal{F}$ such that (7) holds, we use the *chaining* technique (Pollard, 2012; Li et al., 2001). In Lemma 5 (see Section B of the Supplementary Materials), we show that, given independent samples $x_1, x_2, \ldots, x_{2m}$,

$$\mathbb{P}\left[\exists f \in \mathcal{F} : \left|\frac{1}{m}\sum_{i=1}^{m}\sigma_i(f(x_i) - f(x_{i+m}))\right| > \epsilon\right]$$

$$\leq 4\left(16e^2\right)^{\mathrm{Pdim}(\mathcal{F})} e^{-\frac{\epsilon^2 m}{200\frac{1}{2m}\sum_{i=1}^{2m} s(x_i)^2}}.$$

The key difficulty in proving Lemma 5 is that the functions $f \in \mathcal{F}$ are not bounded uniformly in $[0, 1]$. To this end, we provide in Lemma 4 a novel result that bounds the size of $\epsilon$-packings of $\mathcal{F}$ if the functions $f \in \mathcal{F}$ are bounded in expectation. Based on Lemma 5, we then prove the main claim of Theorem 5. □

## 6. Analysis for $k$-Means

In order to apply Theorem 5 to $k$-Means clustering, we require a suitable family $\mathcal{F}$, an upper bound $s(x)$ and a bound on $\mathbb{E}_P\left[s(x)^2\right]$. We provide this in Lemma 1 and defer bounding $\frac{1}{2m} \sum_{i=1}^{2m} s(x_i)^2$ to the proofs of Theorems 2-4.

**Lemma 1** ($k$-Means). *Let $k \in \mathbb{N}$. Let $P$ be any distribution on $\mathbb{R}^d$ with $\mu = \mathbb{E}_P[x]$, $\sigma^2 = \mathbb{E}_P\left[d(x, \mu)^2\right] \in (0, \infty)$ and bounded kurtosis $\hat{M}_4$. For any $x \in \mathbb{R}^d$ and any $Q \in \mathbb{R}^{d \times k}$, define*

$$f_Q(x) = \frac{d(x, Q)^2}{\frac{1}{2}\sigma^2 + \frac{1}{2}\mathbb{E}_P\left[d(x, Q)^2\right]} \qquad (8)$$

*as well as the function family $\mathcal{F} = \{f_Q(\cdot) \mid Q \in \mathbb{R}^{d \times k}\}$. Let*

$$s(x) = \frac{4 \, d(x, \mu)^2}{\sigma^2} + 8.$$

*We then have*

$$\mathrm{Pdim}(\mathcal{F}) \leq 6k(d + 4) \log 6k, \qquad (9)$$

$$f_Q(x) \leq s(x) \quad \forall x \in \mathbb{R}^d \qquad (10)$$

*for any $x \in \mathbb{R}^d$ and $Q \in \mathbb{R}^{d \times k}$ and*

$$\mathbb{E}_P\left[s(x)^2\right] = 128 + 16\hat{M}_4. \qquad (11)$$

The proof of Lemma 1 is provided in Section C of the Supplementary Materials. The definition of $f_Q(x)$ in (8) is motivated as follows: If we use Theorem 5 to guarantee

$$\left|\sum_{i=1}^m f(x_i) - \mathbb{E}_P\left[f(x)\right]\right| \leq \epsilon \quad \forall f \in \mathcal{F}, \qquad (12)$$

then this implies

$$\left|\phi_\mathcal{X}(Q) - \mathbb{E}_P\left[d(x, Q)^2\right]\right| \leq \frac{\epsilon}{2}\sigma^2 + \frac{\epsilon}{2}\mathbb{E}_P\left[d(x, Q)^2\right] \qquad (13)$$

as is required by Theorems 2-4. Lemma 1 further shows that the expectation of $s(x)^2$ is bounded if the kurtosis of $P$ is bounded. This is the reason why a bounded kurtosis is the weakest assumption on $P$ that we require in Section 4.

We now proceed to prove Theorems 2-4 by applying Theorem 5 and examining the tail behavior of $\frac{1}{2m}\sum_{i=1}^{2m} s(x_i)^2$.

### 6.1. Proof of Theorem 1 (kurtosis bound)

The bound based on the kurtosis follows easily from Markov's inequality.

*Proof.* We consider the choice $t = 4\left(128 + 16\hat{M}_4\right)/\delta$. By Markov's inequality and linearity of expectation, we then have that

$$\mathbb{P}\left[\frac{1}{2m}\sum_{i=1}^{2m} s(x_i)^2 > t\right] \leq \frac{\mathbb{E}\left[s(x)^2\right]}{t} = \frac{\delta}{4}.$$

Furthermore, $\mathbb{E}_P\left[s(x)^2\right] \leq t$ by Lemma 1. Hence, we may apply Theorem 5 to obtain that for

$$m \geq \frac{12800\left(8 + \hat{M}_4\right)}{\epsilon^2 \delta}\left(3 + 30k(d+4)\log 6k + \log \frac{1}{\delta}\right),$$

it holds with probability at least $1 - \delta$ that

$$\left|\frac{1}{m}\sum_{i=1}^m f(x_i) - \mathbb{E}\left[f(x)\right]\right| \leq \epsilon \quad \forall f \in \mathcal{F}.$$

This implies the main claim and thus concludes the proof. □

### 6.2. Proof of Theorem 2 (higher order moment bound)

We prove the result by bounding the higher moments of $\frac{1}{2m}\sum_{i=1}^{2m} s(x_i)^2$ using the Marcinkiewicz-Zygmund inequality and subsequently applying Markov's inequality.

*Proof.* Hoelder's inequality implies

$$\hat{M}_4 = \frac{\mathbb{E}_P\left[d(x,\mu)^4\right]}{\sigma^2} \leq \frac{\mathbb{E}_P\left[d(x,\mu)^p\right]^{\frac{4}{p}}}{\sigma^4} \leq \hat{M}_p^{\frac{4}{p}}$$

Hence, by Lemma 1 we have that $\mathbb{E}_P\left[s(x)^2\right] \leq 128 + 16\hat{M}_p^{\frac{4}{p}}$ Since $s(x)^2 \geq 0$ for all $x \in \mathbb{R}^d$, we have

$$\left|s(x)^2 - \mathbb{E}_P\left[s(x)^2\right]\right| \leq \max\left(s(x)^2, \mathbb{E}_P\left[s(x)^2\right]\right)$$

$$\leq \max\left(s(x)^2, 128 + 16\hat{M}_p^{\frac{4}{p}}\right)$$

$$\leq 128$$

$$+ 16 \max\left(\hat{M}_p^{\frac{4}{p}}, 2\frac{d(x,\mu)^4}{\sigma^4}\right). \qquad (14)$$

This implies that

$$\mathbb{E}_P\left[|s(x)^2 - \mathbb{E}_P\left[s(x)^2\right]|^{\frac{p}{4}}\right]$$
$$\leq 256^{\frac{p}{4}} + 32^{\frac{p}{4}} \max\left(\hat{M}_p, 2^{\frac{p}{4}} \frac{\mathbb{E}_P\left[d(x,\mu)^p\right]}{\sigma^p}\right) \quad (15)$$
$$\leq 256^{\frac{p}{4}} + 32^{\frac{p}{4}} \max\left(\hat{M}_p, 2^{\frac{p}{4}} \hat{M}_p\right)$$
$$\leq 256^{\frac{p}{4}} + 64^{\frac{p}{4}} \hat{M}_p.$$

We apply a variant of the Marcinkiewicz-Zygmund inequality (Ren & Liang, 2001) to the zero-mean random variable $s(x)^2 - \mathbb{E}_P\left[s(x)^2\right]$ to obtain

$$\mathbb{E}_P\left[\left|\frac{1}{2m}\sum_{i=1}^{2m}\left(s(x_i)^2 - \mathbb{E}_P\left[s(x)^2\right]\right)\right|^{\frac{p}{4}}\right]$$
$$\leq \left(\frac{p-4}{4\sqrt{2m}}\right)^{\frac{p}{4}} \mathbb{E}_P\left[|s(x)^2 - \mathbb{E}_P\left[s(x)^2\right]|^{\frac{p}{4}}\right] \quad (16)$$
$$\leq \left(\frac{p-4}{4\sqrt{2m}}\right)^{\frac{p}{4}} \left(256^{\frac{p}{4}} + 64^{\frac{p}{4}} \hat{M}_p\right)$$

For $u > 0$, the Markov inequality implies

$$\mathbb{P}\left[\left|\frac{1}{2m}\sum_{i=1}^{2m}\left(s(x_i)^2 - \mathbb{E}_P\left[s(x)^2\right]\right)\right| > u\right]$$
$$\leq \left(\frac{p-4}{4u\sqrt{2m}}\right)^{\frac{p}{4}} \left(256^{\frac{p}{4}} + 64^{\frac{p}{4}} \hat{M}_p\right) \quad (17)$$
$$\leq 2\left(\frac{p-4}{u\sqrt{2m}}\left(64 + 16\hat{M}_p^{\frac{4}{p}}\right)\right)^{\frac{p}{4}}$$

For $u = (p-4)\left(64 + 16\hat{M}_p^{\frac{4}{p}}\right)$, we thus have

$$\mathbb{P}\left[\left|\frac{1}{2m}\sum_{i=1}^{2m}\left(s(x_i)^2 - \mathbb{E}_P\left[s(x)^2\right]\right)\right| > u\right] \leq 2m^{-\frac{p}{8}} \quad (18)$$

Since $m \geq \left(\frac{8}{\delta}\right)^{\frac{8}{p}}$, this implies

$$\mathbb{P}\left[\left|\frac{1}{2m}\sum_{i=1}^{2m}\left(s(x_i)^2 - \mathbb{E}_P\left[s(x)^2\right]\right)\right| > u\right] \leq \frac{\delta}{4} \quad (19)$$

It holds that

$$u + \mathbb{E}_P\left[s(x)^2\right] = (p-4)\left(64 + 16\hat{M}_p^{\frac{4}{p}}\right) + 128 + 16\hat{M}_p^{\frac{4}{p}}$$
$$\leq p\left(64 + 16\hat{M}_p^{\frac{4}{p}}\right) \quad (20)$$

We set $t = p\left(64 + 16\hat{M}_p^{\frac{4}{p}}\right)$ and thus have

$$\mathbb{P}\left[\frac{1}{2m}\sum_{i=1}^{2m}s(x_i)^2 > t\right] \leq \frac{\delta}{4} \quad (21)$$

In combination with $\mathbb{E}_P\left[s(x)^2\right] \leq t$ by Lemma 1, we may thus apply Theorem 5. Since $m \geq \frac{3200m_1}{\epsilon^2}$ with

$$m_1 = p\left(4 + \hat{M}_p^{\frac{4}{p}}\right)\left(3 + 30k(d+4)\log 6k + \log\frac{1}{\delta}\right)$$

it holds with probability at least $1 - \delta$ that

$$\left|\frac{1}{m}\sum_{i=1}^{m}f(x_i) - \mathbb{E}\left[f(x)\right]\right| \leq \epsilon \quad \forall f \in \mathcal{F}.$$

This implies the main claim and thus concludes the proof. $\square$

### 6.3. Proof of Theorem 3 (subgaussianity)

Under subgaussianity, all moments of $d(x, \mu)$ are bounded. We show the result by optimizing over $p$ in Theorem 2.

*Proof.* For $p \in \{4, 8, \ldots, \infty\}$, we have

$$\hat{M}_p = \mathbb{E}_P\left[\left|\frac{d(x,\mu)}{\sigma}\right|^p\right]$$
$$= \int_0^\infty \mathbb{P}\left[\frac{d(x,\mu)}{\sigma} > u^{\frac{1}{p}}\right] du \quad (22)$$
$$\leq \int_0^\infty a \exp\left(-\frac{u^{\frac{2}{p}}}{\sqrt{b}}\right) du.$$

Let $u(t) = b^{\frac{p}{4}} t^{\frac{p}{2}}$ which implies $du/dt = b^{\frac{p}{4}} \frac{p}{2} t^{\frac{p}{2}-1}$. Hence,

$$\hat{M}_p \leq \frac{ab^{\frac{p}{4}}p}{2}\int_0^\infty e^{-t}t^{\frac{p}{2}-1}dt.$$

By the definition of the gamma function and since $p$ is even, we have

$$\int_0^\infty e^{-t}t^{\frac{p}{2}-1}dt = \Gamma\left(\frac{p}{2}\right) = \left(\frac{p}{2}-1\right)! \leq \left(\frac{p}{2}\right)^{\frac{p}{2}-1}$$

Hence, for $p \in \{4, 8, \ldots, \infty\}$, we have

$$\hat{M}_p^{\frac{4}{p}} \leq \frac{1}{4}a^{\frac{4}{p}}bp^2 \leq \frac{1}{4}abp^2.$$

Let $p_* = 4\left\lceil \frac{5}{4} + \frac{3}{4}\log\frac{1}{\delta} \right\rceil$ which implies

$$p_* \geq 5 + 3\log\frac{1}{\delta} \geq \frac{8}{\log 48}\log\frac{8}{\delta}$$

Uniform Deviation Bounds for Unbounded Loss Functions like $k$-Meansand thus $\left(\frac{8}{\delta}\right)^{\frac{8}{p_*}} \leq 48$. We instantiate Theorem 2 with the $p_*$th-order bound $\hat{M}_{p_*}$ of $P$. Since $\left(\frac{8}{\delta}\right)^{\frac{8}{p_*}} \leq 48$, the minimum sample size is thus

$$\frac{3200 p_*}{\epsilon^2}\left(4 + \frac{abp_*{}^2}{4}\right)\left(3 + 30k(d+4)\log 6k + \log\frac{1}{\delta}\right).$$

The main claim finally holds since $p_* \leq p = 9 + 3\log\frac{1}{\delta}$. □

### 6.4. Proof of Theorem 4 (bounded support)

*Proof.* Let $t = 128 + 64R^4/\sigma^4$. Since the support of $P$ is bounded, we have $s(x) \leq t$ for all $x \in \mathbb{R}^d$. This implies that $\mathbb{E}_P\left[s(x)^2\right] \leq t$ and that $\frac{1}{2m}\sum_{i=1}^{2m} s(x_i)^2 \leq t$ almost surely. The result then follows from Theorem 5. □

## 7. Conclusion

We have presented a framework to uniformly approximate the expected value of unbounded functions on an empirical sample. With this framework we are able to provide theoretical guarantees for empirical risk minimization in $k$-Means clustering if the kurtosis of the underlying distribution is bounded. We have obtained state-of-the art bounds on the number of required samples to achieve a given uniform approximation error. If the underlying distribution fulfills stronger assumptions such as bounded higher moment, subgaussianity or bounded support, then we obtain progressively better bounds.

## References

Antos, András, Gyorfi, L, and Gyorgy, Andras. Individual convergence rates in empirical vector quantizer design. *IEEE Transactions on Information Theory*, 51(11):4013–4022, 2005.

Bartlett, Peter L, Linder, Tamás, and Lugosi, Gábor. The minimax distortion redundancy in empirical quantizer design. *IEEE Transactions on Information theory*, 44(5):1802–1813, 1998.

Ben-David, Shai. A framework for statistical clustering with constant time approximation algorithms for k-median and k-means clustering. *Machine Learning*, 66(2-3):243–257, 2007.

Ben-David, Shai, Von Luxburg, Ulrike, and Pál, Dávid. A sober look at clustering stability. In *International Conference on Computational Learning Theory*, pp. 5–19. Springer, 2006.

Har-Peled, Sariel. *Geometric approximation algorithms*, volume 173. American mathematical society Boston, 2011.

Haussler, David. Decision theoretic generalizations of the pac model for neural net and other learning applications. *Information and computation*, 100(1):78–150, 1992.

Hoeffding, Wassily. Probability inequalities for sums of bounded random variables. *Journal of the American statistical association*, 58(301):13–30, 1963.

Levrard, Clément et al. Fast rates for empirical vector quantization. *Electronic Journal of Statistics*, 7:1716–1746, 2013.

Li, Yi, Long, Philip M, and Srinivasan, Aravind. Improved bounds on the sample complexity of learning. *Journal of Computer and System Sciences*, 62(3):516–527, 2001.

Linder, Tamás, Lugosi, Gábor, and Zeger, Kenneth. Rates of convergence in the source coding theorem, in empirical quantizer design, and in universal lossy source coding. *IEEE Transactions on Information Theory*, 40(6):1728–1740, 1994.

Moors, JJA. The meaning of kurtosis: Darlington reexamined. *The American Statistician*, 40(4):283–284, 1986.

Pollard, David. *Convergence of stochastic processes*. Springer Science & Business Media, 2012.

Pollard, David et al. Strong consistency of $k$-means clustering. *The Annals of Statistics*, 9(1):135–140, 1981.

Rakhlin, Alexander and Caponnetto, Andrea. Stability of k-means clustering. *Advances in neural information processing systems*, 19:1121, 2007.

Ren, Yao-Feng and Liang, Han-Ying. On the best constant in marcinkiewicz–zygmund inequality. *Statistics & probability letters*, 53(3):227–233, 2001.

Sauer, Norbert. On the density of families of sets. *Journal of Combinatorial Theory, Series A*, 13(1):145–147, 1972.

Shamir, Ohad and Tishby, Naftali. Cluster stability for finite samples. In *NIPS*, pp. 1297–1304, 2007.

Shamir, Ohad and Tishby, Naftali. Model selection and stability in k-means clustering. In *COLT*, pp. 367–378. Citeseer, 2008.

Telgarsky, Matus J and Dasgupta, Sanjoy. Moment-based uniform deviation bounds for $k$-means and friends. In *Advances in Neural Information Processing Systems*, pp. 2940–2948, 2013.

Vapnik, VN and Chervonenkis, A Ya. On the uniform convergence of relative frequencies of events to their probabilities. *Theory of Probability & Its Applications*, 16(2):264–280, 1971.

## A. Auxiliary lemmas

For the following proofs we require two auxiliary lemmas.

**Lemma 2.** *Let $x > 0$ and $a > 0$. If*

$$x \leq a \log x \tag{23}$$

*then it holds that*

$$x \leq 2a \log 2a. \tag{24}$$

*Proof.* Since $x > 0$, we have $\sqrt{x} > 0$ and thus $\log \sqrt{x} \leq \sqrt{x}$. Together with (23), this implies

$$x \leq a \log x = 2a \log \sqrt{x} \leq 2a\sqrt{x},$$

and thus

$$x \leq 4a^2.$$

We show the result by contradiction. Suppose that

$$x > 2a \log 2a.$$

Together with (23), this implies

$$2a \log 2a < a \log x$$

which in turn leads to the contradiction

$$x > 4a^2.$$

This concludes the proof since (24) must hold. □

**Lemma 3.** *For $n \in \mathbb{N}$, define*

$$S_n = \sum_{j=1}^{n} \sqrt{\frac{j}{2^j}}$$

*Then,*

$$\lim_{n \to \infty} S_n \leq 5.$$

*Proof.* Subtracting

$$\sqrt{\frac{1}{2}} S_n = \sum_{j=1}^{n} \sqrt{\frac{j}{2^{j+1}}} = \sum_{j=2}^{n} \sqrt{\frac{j-1}{2^j}}$$

from $S_n$ yields

$$\left(1 - \sqrt{\frac{1}{2}}\right) S_n = \sum_{j=1}^{n} \frac{\sqrt{j} - \sqrt{j-1}}{2^{j/2}}$$

$$= \sqrt{\frac{1}{2}} + \sum_{j=2}^{n} \frac{\sqrt{j} - \sqrt{j-1}}{2^{j/2}}.$$

For $j \geq 2$, we have $\sqrt{j} - \sqrt{j-1} \leq \sqrt{2} - 1$ and hence

$$\left(1 - \sqrt{\frac{1}{2}}\right) S_n \leq \sqrt{\frac{1}{2}} + \left(\sqrt{2} - 1\right) \sum_{j=2}^{n} \sqrt{\frac{1}{2}}^j$$

$$= \sqrt{\frac{1}{2}} + \frac{\sqrt{2}-1}{2} \sum_{j=2}^{n} \sqrt{\frac{1}{2}}^{j-2}$$

$$= \sqrt{\frac{1}{2}} + \frac{\sqrt{2}-1}{2} \underbrace{\sum_{j=0}^{n} \sqrt{\frac{1}{2}}^j}_{(*)}.$$

The term $(*)$ is a geometric series and hence $\lim_{n \to \infty} \sum_{j=0}^{n} \sqrt{\frac{1}{2}}^j = \left(1 - \sqrt{\frac{1}{2}}\right)^{-1}$. This implies

$$\lim_{n \to \infty} S_n \leq \left[\sqrt{\frac{1}{2}} + \frac{\sqrt{2}-1}{2\left(1-\sqrt{\frac{1}{2}}\right)}\right] \left(1 - \sqrt{\frac{1}{2}}\right)^{-1}$$

$$= \frac{1}{\sqrt{2}\left(1-\sqrt{\frac{1}{2}}\right)} + \frac{\sqrt{2}\left(1-\sqrt{\frac{1}{2}}\right)}{2\left(1-\sqrt{\frac{1}{2}}\right)^2}$$

$$= \frac{2}{\sqrt{2}-1}$$

$$= \frac{2}{\sqrt{2}-1} \frac{\sqrt{2}+1}{\sqrt{2}+1}$$

$$= 2 + 2\sqrt{2}$$

$$\leq 5$$

as desired. □

## B. Proof of Theorem 5

We first show two results, Lemma 4 and 5 and then use them to prove Theorem 5.

**Definition 2.** *Let $\mathcal{F}$ be a family of functions from $\mathcal{X}$ to $\mathbb{R}_{\geq 0}$ and $Q$ an arbitrary measure on $\mathcal{X}$. For any $f, g \in \mathcal{F}$, we define the distance function*

$$d_{L^1(Q)}(f, g) = \int_{\mathcal{X}} |f(x) - g(x)| \, dQ(x).$$

*For any $f \in \mathcal{F}$ and $A \subseteq \mathcal{F}$, we further define*

$$d_{L^1(Q)}(f, A) = \min_{g \in A} d_{L^1(Q)}(f, g).$$

**Definition 3.** *For $\epsilon > 0$, a set $A \subseteq B$ is an $\epsilon$-packing of $B$ with respect to some metric $d$ if for any two distinct $x, y \in A$, $d(x, y) > \epsilon$. The cardinality of the largest $\epsilon$-packing of $B$ with respect to $d$ is denoted by $\mathcal{M}(\epsilon, B, d)$.*

**Lemma 4** ($\epsilon$-packing). *Let $\mathcal{F}$ be a family of functions from $\mathcal{X}$ to $\mathbb{R}_{\geq 0}$ with $\mathrm{Pdim}(F) = d$. For all $x \in \mathcal{X}$, let $s(x) = \sup_{f \in \mathcal{F}} f(x)$. Let $Q$ be an arbitrary measure on $\mathcal{X}$ with $0 < \mathbb{E}_Q\left[s(x)\right] < \infty$. Then, for all $0 < \epsilon \leq \mathbb{E}_Q\left[s(x)\right]$,*

$$\mathcal{M}\left(\epsilon, \mathcal{F}, \mathrm{d}_{L^1(Q)}\right) \leq 8 \left(\frac{2e\mathbb{E}_Q\left[s(x)\right]}{\epsilon}\right)^{2d}.$$

*Proof.* Our proof is similar to the proof of Theorem 6 in Haussler (1992). The difference is that we consider a function family $\mathcal{F}$ that is not uniformly bounded but only bounded in expectation. The key idea is to construct a random sample and to use the expected number of dichotomies on that set to bound the size of an $\epsilon$-packing by the pseudo-dimension.

Noting that by definition $s(x) \geq 0$ and $\mathbb{E}_Q\left[s(x)\right] < \infty$, we define the probability measure $\tilde{Q}$ on $\mathcal{X}$ using the Radon-Nikodym derivative

$$\frac{d\tilde{Q}(x)}{dQ(x)} = \frac{s(x)}{\mathbb{E}_Q\left[s(x)\right]}, \quad \forall x \in \mathcal{X}.$$

Let $\vec{x} = (x_1, x_2, \ldots, x_m)$ be a random vector in $\mathcal{X}^m$, where each $x_i$ is drawn independently at random from $\tilde{Q}$. Given $\vec{x}$, let $\vec{r} = (r_1, r_2, \ldots, r_m)$ be a random vector, where each $r_i$ is drawn independently at random from a uniform distribution on $[0, s(x_i)]$.

For any $f \in \mathcal{F}$, we denote the restriction $(f(x_1), \ldots, f(x_m))$ by $f_{\vec{x}}$ and set $\mathcal{F}_{\vec{x}} = \{f_{\vec{x}} \mid f \in \mathcal{F}\}$. For any vector $\vec{z} \in \mathbb{R}^m$, we define

$$\mathrm{sign}(\vec{z}) = (\mathrm{sign}(z_1), \ldots, \mathrm{sign}(z_m)).$$

The set of dichotomies induced by $\vec{r}$ on $F_{\vec{x}}$ is given by

$$\mathrm{sign}(\mathcal{F}_{\vec{x}} - \vec{r}) = \{\mathrm{sign}(f_{\vec{x}} - \vec{r}) \mid f \in \mathcal{F}\}.$$

For $m \geq d$, Sauer's Lemma (Sauer, 1972; Vapnik & Chervonenkis, 1971) bounds the size of this set by

$$|\mathrm{sign}(\mathcal{F}_{\vec{x}} - \vec{r})| \leq (em/d)^d,$$

for all $\vec{x} \in \mathcal{X}^m$ and $\vec{r} \in \mathbb{R}^m$. Hence, the expected number of dichotomies is also bounded, i.e.

$$\mathbb{E}\left[|\mathrm{sign}(\mathcal{F}_{\vec{x}} - \vec{r})|\right] \leq (em/d)^d. \tag{25}$$

Let $\mathcal{G}$ be a $\epsilon$-separated subset of $\mathcal{F}$ with respect to $\mathrm{d}_{L^1(Q)}$ with $|\mathcal{G}| = \mathcal{M}\left(\epsilon, \mathcal{F}, \mathrm{d}_{L^1(Q)}\right)$. By definition, for any two distinct $f, g \in \mathcal{G}$, we have

$$\int_{\mathcal{X}} |f(x) - g(x)| \, dQ(x) > \epsilon.$$

Consider the set $\mathcal{X}_0 = \{x \in \mathcal{X} : s(x) = 0\}$ and define $\mathcal{X}_{>0} = \mathcal{X} \setminus \mathcal{X}_0$. By definition of $\tilde{Q}$, $\mathcal{X}_0$ is zero set of $\tilde{Q}$ and, since $f(x), g(x) \in [0, s(x)]$ for all $x \in \mathcal{X}$, we have $\int_{\mathcal{X}_0} |f(x) - g(x)| \, dQ(x) = 0$.

For any two distinct $f, g \in \mathcal{G}$, we thus have for all $i = 1, \ldots, m$

$$\begin{aligned}
&\mathbb{P}\left[\mathrm{sign}(f(x_i) - r_i) \neq \mathrm{sign}(g(x_i) - r_i)\right] \\
&= \int_{\mathcal{X}} \int_0^{|f(x_i) - g(x_i)|} \frac{1}{s(x)} dr d\tilde{Q}(x) \\
&= \int_{\mathcal{X}_{>0}} \int_0^{|f(x_i) - g(x_i)|} \frac{1}{s(x)} dr d\tilde{Q}(x) \\
&= \int_{\mathcal{X}_{>0}} \frac{|f(x_i) - g(x_i)|}{s(x)} d\tilde{Q}(x) \\
&= \int_{\mathcal{X}_{>0}} \frac{|f(x_i) - g(x_i)|}{\mathbb{E}_Q\left[s(x)\right]} dQ(x) \\
&= \int_{\mathcal{X}} \frac{|f(x_i) - g(x_i)|}{\mathbb{E}_Q\left[s(x)\right]} dQ(x) \\
&> \frac{\epsilon}{\mathbb{E}_Q\left[s(x)\right]}
\end{aligned}$$

This allows us to bound the probability that two distinct $f, g \in \mathcal{G}$ produce the same dichotomy on all $m$ samples, i.e.

$$\begin{aligned}
&\mathbb{P}\left[\mathrm{sign}(f_{\vec{x}} - \vec{r}) = \mathrm{sign}(g_{\vec{x}} - \vec{r})\right] \\
&= \prod_{i=1}^m \left(1 - \mathbb{P}\left[\mathrm{sign}(f(x_i) - r_i) \neq \mathrm{sign}(g(x_i) - r_i)\right]\right) \\
&\leq \left(1 - \frac{\epsilon}{\mathbb{E}_Q\left[s(x)\right]}\right)^m \leq \exp\left(-\frac{\epsilon m}{\mathbb{E}_Q\left[s(x)\right]}\right).
\end{aligned}$$

Given $\vec{x} \in \mathcal{X}^m$ and $\vec{r} \in \mathbb{R}^m$, let $\mathcal{H}$ be the subset of $\mathcal{G}$ with unique dichotomies, i.e., $\mathcal{H} \subseteq \mathcal{G}$ such that for any $f \in \mathcal{H}$,

$$\mathrm{sign}(f_{\vec{x}} - \vec{r}) \neq \mathrm{sign}(g_{\vec{x}} - \vec{r}),$$

for all $g \in \mathcal{G} \setminus \{f\}$. We then have

$$\begin{aligned}
&\mathbb{P}\left[f \notin \mathcal{H}\right] \\
&= \mathbb{P}\left[\exists g \in \mathcal{G} \setminus \{f\} : \mathrm{sign}(f_{\vec{x}} - \vec{r}) = \mathrm{sign}(g_{\vec{x}} - \vec{r})\right] \\
&\leq |\mathcal{G}| \max_{g \in \mathcal{G} \setminus \{f\}} \mathbb{P}\left[\mathrm{sign}(f_{\vec{x}} - \vec{r}) = \mathrm{sign}(g_{\vec{x}} - \vec{r})\right] \\
&\leq |\mathcal{G}| \cdot \exp\left(-\frac{\epsilon m}{\mathbb{E}_Q\left[s(x)\right]}\right).
\end{aligned}$$

This allows us to bound the expected number of dichotomies from below, i.e.,

$$\begin{aligned}
\mathbb{E}\left[|\text{sign}(\mathcal{F}_{\vec{x}} - \vec{r})|\right] &\geq \mathbb{E}\left[|\text{sign}(\mathcal{G}_{\vec{x}} - \vec{r})|\right] \\
&\geq \mathbb{E}\left[|\text{sign}(\mathcal{H}_{\vec{x}} - \vec{r})|\right] \\
&\geq \mathbb{E}\left[|\mathcal{H}|\right] \\
&= \sum_{f \in \mathcal{G}} (1 - \mathbb{P}\left[f \notin \mathcal{H}\right]) \\
&\geq |\mathcal{G}| \left[1 - |\mathcal{G}| \cdot \exp\left(-\frac{\epsilon m}{\mathbb{E}_Q\left[s(x)\right]}\right)\right].
\end{aligned}$$

Together with (25), we thus have for $m \geq d$

$$\left(\frac{em}{d}\right)^d \geq |\mathcal{G}| \left[1 - |\mathcal{G}| \cdot \exp\left(-\frac{\epsilon m}{\mathbb{E}_Q\left[s(x)\right]}\right)\right]. \quad (26)$$

Consider the case

$$\frac{\mathbb{E}_Q\left[s(x)\right]}{\epsilon} \ln(2|G|) < d.$$

Since $\epsilon \leq \mathbb{E}_Q\left[s(x)\right]$ and $|\mathcal{G}| = \mathcal{M}\left(\epsilon, \mathcal{F}, d_{L^1(Q)}\right)$, we then have

$$\mathcal{M}\left(\epsilon, \mathcal{F}, d_{L^1(Q)}\right) = |\mathcal{G}| \leq \frac{1}{2} e^{d\epsilon / \mathbb{E}_Q[s(x)]} \leq \frac{1}{2} e^d$$

as required to show the result. We hence assume $\mathbb{E}_Q\left[s(x)\right] \ln(2|G|)/\epsilon \geq d$ for the remainder of the proof.

Let $m \geq \mathbb{E}_Q\left[s(x)\right] \ln(2|G|)/\epsilon$ which implies

$$1 - |\mathcal{G}| \cdot \exp\left(-\frac{\epsilon m}{\mathbb{E}_Q\left[s(x)\right]}\right) \geq \frac{1}{2}.$$

Together with (26) and $m \geq d$, it follows that

$$|\mathcal{G}| \leq 2 \left(\frac{e \mathbb{E}_Q\left[s(x)\right]}{\epsilon d} \ln(2|G|)\right)^d$$

and hence

$$\sqrt{|G|} \frac{2^d d^d \sqrt{2|\mathcal{G}|}}{(\ln 2|G|)^d} \leq 2\sqrt{2} \left(\frac{2e \mathbb{E}_Q\left[s(x)\right]}{\epsilon}\right)^d. \quad (27)$$

Since $\ln x \leq x$, we have for $x = (2|G|)^{1/2d}$ that

$$\ln(2|G|)^{1/2d} \leq (2|G|)^{1/2d}$$

which implies

$$\ln(2|G|) \leq 2d(2|G|)^{1/2d}$$

and hence

$$1 \leq \frac{2^d d^d \sqrt{2|\mathcal{G}|}}{(\ln 2|G|)^d}.$$

Together with (27) and $|\mathcal{G}| = \mathcal{M}\left(\epsilon, \mathcal{F}, d_{L^1(Q)}\right)$, we have

$$\mathcal{M}\left(\epsilon, \mathcal{F}, d_{L^1(Q)}\right) = |\mathcal{G}| \leq 8 \left(\frac{2e \mathbb{E}_Q\left[s(x)\right]}{\epsilon}\right)^{2d}$$

as required which concludes the proof.

$\square$

**Lemma 5** (Chaining). *Let $\mathcal{F}$ be a family of functions from $\mathcal{X}$ to $\mathbb{R}_{\geq 0}$ with $\text{Pdim}(F) = d < \infty$. For all $x \in \mathcal{X}$, let $s(x) = \sup_{f \in \mathcal{F}} f(x)$. For $m \geq 200K(2d+1)/\epsilon^2$, let $x_1, \ldots, x_{2m}$ be a subset of $\mathcal{X}$ with $K = \frac{1}{2m} \sum_{i=1}^{2m} s(x_i)^2 < \infty$. For $i = 1, 2, \ldots, m$, let $\sigma_i$ be drawn from $\{-1, 1\}$ uniformly at random. Then, for all $0 < \epsilon \leq 1$,*

$$\mathbb{P}\left[\exists f \in \mathcal{F} : \left|\frac{1}{m} \sum_{i=1}^m \sigma_i(f(x_i) - f(x_{i+m}))\right| > \epsilon\right]$$
$$\leq 4 \left(16 e^2\right)^d e^{-\frac{\epsilon^2 m}{200 K}}.$$

*Proof.* Consider the case $\frac{1}{2m} \sum_{i=1}^{2m} s(x_i) \leq 0$. By definition, we have $s(x_i) \geq f(x_i) \geq 0$ for all $f \in \mathcal{F}$ and $i = 1, \ldots, 2m$. Thus, $f(x_i) = 0$ for all $i = 1, \ldots, 2m$. The claim then follows directly since

$$\sum_{i=1}^m \sigma_i(f(x_i) - f(x_{i+m})) = 0$$

for all $f \in \mathcal{F}$. For the remainder of the proof, we hence only need to consider the case $\frac{1}{2m} \sum_{i=1}^{2m} s(x_i) > 0$.

We define the discrete measure $Q$ by placing an atom at each $x_i$ with weight proportional to $s(x_i)$. More formally,

$$\mathbb{P}_{X \sim Q}[X = x] = \sum_{i=1}^{2m} \frac{s(x_i)}{\sum_{k=1}^{2m} s(x_k)} \mathbf{1}_{\{x_i = x\}}, \quad \forall x \in \mathcal{X}.$$

Since $\sum_{i=1}^{2m} s(x_i)^2 < \infty$ and $\sum_{i=1}^{2m} s(x_i) > 0$, we have

$$\mathbb{E}_Q\left[s(x)\right] = \frac{\sum_{i=1}^{2m} s(x_i)^2}{\sum_{k=1}^{2m} s(x_k)} < \infty. \quad (28)$$

For $j \in \mathbb{N}$, let $\gamma_j = \mathbb{E}_Q\left[s(x)\right]/2^j$. We define a sequence $\mathcal{G}_1, \mathcal{G}_2, \ldots, \mathcal{G}_\infty$ of $\gamma_j$-packings of $\mathcal{F}$ as follows: Let the set $\mathcal{G}_0$ consist of an arbitrary element $f \in \mathcal{F}$. For any $j \in \mathbb{N}$, we initialize $\mathcal{G}_j$ to $\mathcal{G}_{j-1}$. Then, we select a single element $f \in \mathcal{F}$ with $d_{L^1(Q)}(f, \mathcal{G}_j) > \gamma_j$ and add it to $\mathcal{G}_j$. We repeat this until no such element $f \in \mathcal{F}$ with $d_{L^1(Q)}(f, \mathcal{G}_j) > \gamma_j$ is left. By definition, $\mathcal{G}_j$ is an $\gamma_j$-packing of $\mathcal{F}$ with respect to $d_{L^1(Q)}$. Hence, for any $f \in \mathcal{F}$, we have

$$d_{L^1(Q)}(f, \mathcal{G}_j) \leq \gamma_j = \mathbb{E}_Q\left[s(x)\right]/2^j. \quad (29)$$

By Lemma 4, the size of $\mathcal{G}_j$ is bounded by

$$|\mathcal{G}_j| \leq 2(2e2^j)^{2d} = 2^{2d(j+1)+1}e^{2d}. \qquad (30)$$

For each $f \in \mathcal{F}$ and $j \in \mathbb{N}$, we define the closest element in $\mathcal{G}_j$ by

$$\phi_j(f) = \arg\min_{g \in \mathcal{G}_j} \mathrm{d}_{L^1(Q)}(f,g).$$

By (30), $\mathcal{G}_j$ is finite for each $j \in \mathbb{N}$ and the minimum is well-defined.

We construct the following sequence $\mathcal{H}_1, \mathcal{H}_2, \ldots, \mathcal{H}_\infty$: Let $\mathcal{H}_1$ be equal to $\mathcal{G}_1$. For each $j \in \{2, 3, \ldots, \infty\}$, we define

$$\mathcal{H}_j = \{g - \phi_{j-1}(g) : g \in \mathcal{G}_j\}.$$

For all $j \in \mathbb{N}$ and $h \in \mathcal{H}_j$, there is hence a $g_h \in \mathcal{G}_j$ such that $h = g_h(x) - \phi_{j-1}(g_h)$. By (29), we thus have for all $j \in \mathbb{N}$ and $h \in \mathcal{H}_j$

$$\mathbb{E}_Q\left[|h(x)|\right] = \mathrm{d}_{L^1(Q)}(g_h, \mathcal{G}_{j-1}) \leq \gamma_{j-1}. \qquad (31)$$

Furthermore, by (30), we have for all $j \in \mathbb{N}$

$$|\mathcal{H}_j| \leq |\mathcal{G}_j| \leq 2^{2d(j+1)+1}e^{2d}. \qquad (32)$$

The key idea is that intuitively any $f \in \mathcal{F}$ can be additively decomposed into functions from the sequence $\mathcal{H}_1, \mathcal{H}_2, \ldots, \mathcal{H}_\infty$. By definition, for any $j \in \mathbb{N}$, any function $g \in \mathcal{G}_j$ can be rewritten as

$$g = \sum_{k=1}^{j} h_{g,k}$$

where $(h_{g,1}, h_{g,2}, \ldots, h_{g,j})$ are functions in $\mathcal{H}_1 \times \mathcal{H}_2 \times \ldots \times \mathcal{H}_j$. Let $j \to \infty$ and define

$$\mathcal{G} = \bigcup_{j=1}^{\infty} \mathcal{G}_j.$$

Clearly, $\mathcal{G}$ is dense in $\mathcal{F}$ with respect to $\mathrm{d}_{L^1(Q)}$. We claim that, as a consequence,

$$\exists f \in \mathcal{F} : \left|\frac{1}{m}\sum_{i=1}^{m}\sigma_i(f(x_i) - f(x_{i+m}))\right| > \epsilon \qquad (33)$$

if and only if

$$\exists g \in \mathcal{G} : \left|\frac{1}{m}\sum_{i=1}^{m}\sigma_i(g(x_i) - g(x_{i+m}))\right| > \epsilon. \qquad (34)$$

Since $\mathcal{G} \subseteq \mathcal{F}$, we have (34) $\implies$ (33). To show the converse, assume $\exists f \in \mathcal{F}$ such that

$$\left|\frac{1}{m}\sum_{i=1}^{m}\sigma_i(f(x_i) - f(x_{i+m}))\right| = \epsilon + \kappa,$$

for some $\kappa > 0$. By (29) and $j$ sufficiently large, there exists a $g \in \mathcal{G}$ such that

$$\mathrm{d}_{L^1(Q)}(f,g) < \frac{4}{\frac{1}{2m}\sum_{k=1}^{2m}s(x_k)}\kappa^2. \qquad (35)$$

Using the triangle inequality, we have

$$\epsilon + \kappa = \left|\frac{1}{m}\sum_{i=1}^{m}\sigma_i(f(x_i) - f(x_{i+m}))\right|$$

$$\leq \left|\frac{1}{m}\sum_{i=1}^{m}\sigma_i(g(x_i) - g(x_{i+m}))\right|$$

$$+ \frac{1}{m}\sum_{i=1}^{2m}|f(x_i) - g(x_i)|.$$

Using the Cauchy Schwarz inequality, the fact that $f(x), g(x) \in [0, s(x)]$ for all $x \in \mathcal{X}$, as well as the definition of $\mathrm{d}_{L^1(Q)}$ and (35), we may bound

$$\frac{1}{m}\sum_{i=1}^{2m}|f(x_i) - g(x_i)|$$

$$\leq \sqrt{\frac{1}{m^2}\sum_{i=1}^{2m}|f(x_i) - g(x_i)|^2}$$

$$\leq \sqrt{\frac{1}{m^2}\sum_{i=1}^{2m}|f(x_i) - g(x_i)|\,s(x_i)}$$

$$= \sqrt{\frac{\sum_{k=1}^{2m}s(x_k)}{m^2}\sum_{i=1}^{2m}|f(x_i) - g(x_i)|\frac{s(x_i)}{\sum_{k=1}^{2m}s(x_k)}}$$

$$= \sqrt{4\frac{\sum_{k=1}^{2m}s(x_k)}{2m}\mathrm{d}_{L^1(Q)}(f,g)}$$

$$< \kappa.$$

Together with (B), we hence have

$$\left|\frac{1}{m}\sum_{i=1}^{m}\sigma_i(g(x_i) - g(x_{i+m}))\right| > \epsilon.$$

which implies (33) $\implies$ (34) as claimed.

As a consequence, it is sufficient to only consider $\mathcal{G}$ instead of $\mathcal{F}$. More formally,

$$\mathbb{P}\left[\exists f \in \mathcal{F}: \left|\frac{1}{m}\sum_{i=1}^{m}\sigma_i(f(x_i)-f(x_{i+m}))\right| > \epsilon\right]$$
$$= \mathbb{P}\left[\exists g \in \mathcal{G}: \left|\frac{1}{m}\sum_{i=1}^{m}\sigma_i(g(x_i)-g(x_{i+m}))\right| > \epsilon\right]$$
$$\leq \mathbb{P}\left[\exists g \in \mathcal{G}: \left|\frac{1}{m}\sum_{i=1}^{m}\sum_{j=1}^{\infty}\sigma_i(h_{g,j}(x_i)-h_{g,j}(x_{i+m}))\right| > \epsilon\right]$$
$$\leq \mathbb{P}\left[\exists g \in \mathcal{G}: \sum_{j=1}^{\infty}\left|\frac{1}{m}\sum_{i=1}^{m}\sigma_i(h_{g,j}(x_i)-h_{g,j}(x_{i+m}))\right| > \epsilon\right]$$

For $i \in \mathbb{N}$, let $\epsilon_j = \frac{\epsilon}{5}\sqrt{\frac{j}{2^j}}$. In Lemma 3, we show that $\sum_{j=1}^{\infty}\epsilon_j \leq \epsilon$. Suppose it holds that

$$\left|\frac{1}{m}\sum_{i=1}^{m}\sigma_i(h_{g,j}(x_i)-h_{g,j}(x_{i+m}))\right| \leq \epsilon_j$$

for all $g \in \mathcal{G}$ and $j \in \mathbb{N}$. Then, we have that

$$\sum_{j=1}^{\infty}\left|\frac{1}{m}\sum_{i=1}^{m}\sigma_i(h_{g,j}(x_i)-h_{g,j}(x_{i+m}))\right| \leq \sum_{j=1}^{\infty}\epsilon_j \leq \epsilon$$

for all $g \in \mathcal{G}$. Hence, using the union bound, we have

$$\mathbb{P}\left[\exists f \in \mathcal{F}: \left|\frac{1}{m}\sum_{i=1}^{m}\sigma_i(f(x_i)-f(x_{i+m}))\right| > \epsilon\right]$$
$$\leq \sum_{j=1}^{\infty}\mathbb{P}\left[\exists g \in \mathcal{G}: \left|\frac{1}{m}\sum_{i=1}^{m}\sigma_i(h_{g,j}(x_i)-h_{g,j}(x_{i+m}))\right| > \epsilon_j\right]$$
$$= \sum_{j=1}^{\infty}\mathbb{P}\left[\exists h \in \mathcal{H}_j: \left|\frac{1}{m}\sum_{i=1}^{m}\sigma_i(h(x_i)-h(x_{i+m}))\right| > \epsilon_j\right]$$
$$\leq \sum_{j=1}^{\infty}|\mathcal{H}_j|\max_{h \in \mathcal{H}_j}\mathbb{P}\left[\left|\frac{1}{m}\sum_{i=1}^{m}\sigma_i(h(x_i)-h(x_{i+m}))\right| > \epsilon_j\right]$$
(36)

We now use Hoeffding's inequality to bound the probability that

$$\left|\frac{1}{m}\sum_{i=1}^{m}\sigma_i(h(x_i)-h(x_{i+m}))\right| > \epsilon_j,$$

for a single $j \in \mathbb{N}$ and $h \in \mathcal{H}_j$.

For $i \in \{1,2,\ldots,m\}$, consider the random variables

$$X_i = \sigma_i(h(x_i)-h(x_{i+m})).$$

Since $\sigma_i$ are uniformly drawn at random from $\{-1,1\}$, we have $\mathbb{E}[X_i] = 0$ for all $i \in \{1,2,\ldots,m\}$. Furthermore, each $X_i$ is bounded in

$$[a_i, b_i] = [0 \pm (h(x_i)-h(x_{i+m}))]. \quad (37)$$

Since all $X_i$ are independent, we may apply Hoeffding's inequality. By Theorem 2 of Hoeffding (1963), we have

$$\mathbb{P}\left[\left|\frac{1}{m}\sum_{i=1}^{m}X_i - \mathbb{E}[X_i]\right| > \epsilon_j\right] \leq e^{-\frac{2\epsilon_j^2 m}{\frac{1}{m}\sum_{i=1}^{m}(a_i-b_i)^2}}$$

Using (37), we have

$$(a_i-b_i)^2 = 4(h(x_i)-h(x_{i+m}))^2 \leq 4h(x_i)^2+4h(x_{i+m})^2$$

which implies that

$$\frac{1}{m}\sum_{i=1}^{m}(a_i-b_i)^2 \leq 8\frac{1}{2m}\sum_{i=1}^{2m}h(x_i)^2.$$

Using $h(x_i) \in [0, s(x_i)]$ and the definition of $\mathbb{E}_Q[\cdot]$, we have

$$\frac{1}{m}\sum_{i=1}^{m}(a_i-b_i)^2 \leq 8\frac{1}{2m}\sum_{i=1}^{2m}|h(x_i)|s(x_i)$$
$$= 8\left(\frac{1}{2m}\sum_{i=1}^{2m}s(x_i)\right)\mathbb{E}_Q[|h(x)|]$$

By (31) and (28), we thus have

$$\frac{1}{m}\sum_{i=1}^{m}(a_i-b_i)^2 \leq 8\left(\frac{1}{2m}\sum_{i=1}^{2m}s(x_i)\right)\gamma_{j-1}$$
$$= 2^{4-j}\left(\frac{1}{2m}\sum_{i=1}^{2m}s(x_i)\right)\mathbb{E}_Q[s(x)]$$
$$= 2^{4-j}\left(\frac{1}{2m}\sum_{i=1}^{2m}s(x_i)\right)\frac{\sum_{i=1}^{2m}s(x_i)^2}{\sum_{k=1}^{2m}s(x_k)}$$
$$= 2^{4-j}\left(\frac{1}{2m}\sum_{i=1}^{2m}s(x_i)^2\right)$$
$$= 2^{4-j}K$$

Since $\epsilon_i = \frac{\epsilon}{5}\sqrt{\frac{j}{2^j}}$ this implies

$$-\frac{2\epsilon_j^2 m}{\frac{1}{m}\sum_{i=1}^{m}(a_i-b_i)^2} \leq -\frac{\epsilon^2 mj}{200K}$$

Hence, for any $j \in \mathbb{N}$ and any $h \in \mathcal{H}_j$

$$\mathbb{P}\left[\left|\frac{1}{m}\sum_{i=1}^{m}\sigma_i(h(x_i)-h(x_{i+m}))\right| > \epsilon_j\right] \leq e^{-\frac{\epsilon^2 mj}{200K}}.$$

Together with (32), this allows us to bound (36), i.e.,

$$\mathbb{P}\left[\exists f \in \mathcal{F} : \left|\frac{1}{m}\sum_{i=1}^{m}\sigma_i(f(x_i) - f(x_{i+m}))\right| > \epsilon\right]$$
$$\leq \sum_{j=1}^{\infty}|\mathcal{H}_j|e^{-\frac{\epsilon^2 mj}{200K}}$$
$$\leq \sum_{j=1}^{\infty}2^{2d(j-1)+4d+1}e^{2d}e^{-\frac{\epsilon^2 mj}{200K}}$$
$$= 2^{4d+1}e^{2d}e^{-\frac{\epsilon^2 m}{200K}}\sum_{j=1}^{\infty}\left(4^d e^{-\frac{\epsilon^2 m}{200K}}\right)^{j-1}$$
$$= 2^{4d+1}e^{2d}e^{-\frac{\epsilon^2 m}{200K}}\sum_{j=0}^{\infty}\left(4^d e^{-\frac{\epsilon^2 m}{200K}}\right)^{j}.$$

By assumption in the main claim, we have $m \geq 200K(2d+1)/\epsilon^2$ and hence

$$0 \leq 4^d e^{-\frac{\epsilon^2 m}{200K}} \leq \frac{1}{2}.$$

This implies

$$\sum_{j=0}^{\infty}\left(4^d e^{-\frac{\epsilon^2 m}{100K}}\right)^{j} \leq \sum_{j=0}^{\infty}\frac{1}{2^j} = 2$$

and hence

$$\mathbb{P}\left[\exists f \in \mathcal{F} : \left|\frac{1}{m}\sum_{i=1}^{m}\sigma_i(f(x_i) - f(x_{i+m}))\right| > \epsilon\right]$$
$$\leq 4\left(16e^2\right)^d e^{-\frac{\epsilon^2 m}{200K}}$$

which concludes the proof. $\square$

With these results we are able to prove Theorem 5.

*Proof of Theorem 5.* Our goal is to upper bound the probability of the event

$$A = \left\{\exists f \in \mathcal{F} : \left|\frac{1}{m}\sum_{i=1}^{m}f(x_i) - \mathbb{E}\left[f(x)\right]\right| > \epsilon\right\}$$

by $\delta$, i.e., to prove $\mathbb{P}\left[A\right] \leq \delta$. Consider the event

$$B = \left\{\exists f \in \mathcal{F} : \left|\frac{1}{m}\sum_{i=1}^{m}f(x_i) - \mathbb{E}\left[f(x)\right]\right| > \epsilon \right.$$
$$\left. \cap \left|\frac{1}{m}\sum_{i=m+1}^{2m}f(x_i) - \mathbb{E}\left[f(x)\right]\right| \leq \epsilon/2\right\}.$$

and assume that the event $A$ holds, i.e., there exists a $f' \in \mathcal{F}$ such that

$$\left|\frac{1}{m}\sum_{i=1}^{m}f'(x_i) - \mathbb{E}\left[f'(x)\right]\right| > \epsilon.$$

For any $f \in \mathcal{F}$, Markov's inequality in combination with Jensen's inequality implies

$$\mathbb{P}\left[\left|\frac{1}{m}\sum_{i=m+1}^{2m}f(x_i) - \mathbb{E}\left[f(x)\right]\right| > \frac{\epsilon}{2}\right]$$
$$\leq \frac{4 \cdot \mathbb{E}\left[\left|\frac{1}{m}\sum_{i=m+1}^{2m}f(x_i) - \mathbb{E}\left[f(x)\right]\right|^2\right]}{\epsilon^2}$$
$$\leq \frac{4 \cdot \mathbb{E}\left[\left|\frac{1}{m}\sum_{i=m+1}^{2m}f(x_i)\right|^2\right]}{\epsilon^2}$$
$$\leq \frac{4 \cdot \mathbb{E}\left[\frac{1}{m}\sum_{i=m+1}^{2m}|f(x_i)|^2\right]}{m\epsilon^2}$$
$$= \frac{4 \cdot \mathbb{E}\left[s(x)^2\right]}{m\epsilon^2}.$$

Together with

$$m \geq \frac{200t}{\epsilon^2}\left(3 + 5d + \log\frac{1}{\delta}\right) \geq \frac{8t}{\epsilon^2}$$

this implies that

$$\mathbb{P}\left[\left|\frac{1}{m}\sum_{i=m+1}^{2m}f'(x_i) - \mathbb{E}\left[f'(x)\right]\right| \leq \epsilon/2\right] \geq \frac{1}{2}$$

since $\mathbb{E}_P\left[s(x)^2\right] \leq t$ by (5). Thus, given $A$, the event $B$ holds with probability at least $1/2$, i.e.,

$$\mathbb{P}\left[B \mid A\right] \leq 1/2.$$

Since

$$\mathbb{P}\left[B\right] = \mathbb{P}\left[A \cap B\right] = \mathbb{P}\left[B \mid A\right]\mathbb{P}\left[A\right],$$

we thus have

$$\mathbb{P}\left[A\right] \leq 2 \cdot \mathbb{P}\left[B\right]. \quad (38)$$

We consider the event

$$C = \left\{\exists f \in \mathcal{F} : \left|\frac{1}{m}\sum_{i=1}^{m}(f(x_i) - f(x_{i+m}))\right| > \epsilon/2\right\}$$

and note that, if $B$ holds, then there exists a $f' \in \mathcal{F}$ such that

$$\epsilon < \left| \frac{1}{m} \sum_{i=1}^{m} f'(x_i) - \mathbb{E}\left[ f'(x) \right] \right|$$

$$\leq \left| \frac{1}{m} \sum_{i=1}^{m} f'(x_i) - \frac{1}{m} \sum_{i=m+1}^{2m} f'(x_i) \right|$$

$$+ \underbrace{\left| \frac{1}{m} \sum_{i=m+1}^{2m} f'(x_i) - \mathbb{E}\left[ f'(x) \right] \right|}_{\leq \frac{\epsilon}{2}}$$

which implies that there exists a $f' \in \mathcal{F}$ with

$$\left| \frac{1}{m} \sum_{i=1}^{m} (f'(x_i) - f'(x_{i+m})) \right| > \epsilon/2 \, .$$

Hence, $B \subseteq C$ which in combination with (38) implies that

$$\mathbb{P}\left[ A \right] \leq 2 \cdot \mathbb{P}\left[ B \right] \leq 2 \cdot \mathbb{P}\left[ C \right]. \tag{39}$$

Let $\vec{\sigma} = (\sigma_1, \sigma_2, \ldots, \sigma_m)$ be a random vector where each $\sigma_i$ is sampled independently at random from a uniform distribution on $\{-1, 1\}$. We define the event

$$D = \left\{ f \in \mathcal{F} : \left| \frac{1}{m} \sum_{i=1}^{m} \sigma_i \left( f(x_i) - f(x_{i+m}) \right) \right| > \epsilon/2 \right\}.$$

In essence, $\sigma_i$ randomly permutes $x_i$ and $x_{m+i}$ for any $i \in \{1, \ldots, m\}$. Hence, since all $x_i$ are identically and independently distributed and hence exchangeable, we have

$$\mathbb{P}\left[ C \right] = \mathbb{P}\left[ D \right]. \tag{40}$$

Consider the event

$$E = \left\{ \frac{1}{2m} \sum_{i=1}^{2m} s(x_i)^2 \leq t \right\}$$

and let $\overline{E}$ denote its complement. By (5), we have that

$$\mathbb{P}\left[ \overline{E} \right] \leq \frac{\delta}{4}.$$

Let $\mathbb{E}_{\vec{x}}\left[ \cdot \right]$ denote the expectation with regards to the random vector $\vec{x} = (x_1, x_2, \ldots, x_{2m})$ and $\mathbb{P}_{\vec{\sigma}}\left[ \cdot \right]$ the probability with regards to the random vector $\vec{\sigma}$. By construction, $\vec{\sigma}$ and $\vec{x}$ are independent and the event $E$ only depends on $\vec{x}$ but not on $\vec{\sigma}$. We thus have

$$\mathbb{P}\left[ D \right] = \mathbb{P}\left[ D \cap \overline{E} \right] + \mathbb{P}\left[ D \cap E \right]$$
$$\leq \mathbb{P}\left[ \overline{E} \right] + \mathbb{E}_{\vec{x}}\left[ \mathbb{P}_{\vec{\sigma}}\left[ D \cap E \right] \right]$$
$$\leq \frac{\delta}{4} + \mathbb{E}_{\vec{x}}\left[ \mathbb{P}_{\vec{\sigma}}\left[ D \right] \mathbf{1}_E \right]$$
$$= \frac{\delta}{4} + \mathbb{E}_{\vec{x}}\left[ \mathbb{P}_{\vec{\sigma}}\left[ D | E \right] \mathbf{1}_E \right].$$

Consider any fixed vector $\vec{x} = (x_1, x_2, \ldots, x_{2m})$: If $E$ does not hold, then $\mathbb{P}_{\vec{\sigma}}\left[ D | E \right] \mathbf{1}_E = 0$. Otherwise, $\frac{1}{2m} \sum_{i=1}^{2m} s(x_i)^2 \leq t$ and consequently Lemma 5 with $K = t$ implies that for $m \geq 200t(2d+1)/\epsilon^2$

$$\mathbb{P}_{\vec{\sigma}}\left[ D | E \right] \leq 4 \left( 16e^2 \right)^d e^{-\frac{\epsilon^2 m}{200t}}.$$

As a result, we have for $m \geq 200t(2d+1)/\epsilon^2$

$$\mathbb{P}\left[ D \right] \leq \frac{\delta}{4} + \mathbb{E}_{\vec{x}}\left[ 4 \left( 16e^2 \right)^d e^{-\frac{\epsilon^2 m}{200t}} \mathbf{1}_E \right]$$
$$\leq \frac{\delta}{4} + 4 \left( 16e^2 \right)^d e^{-\frac{\epsilon^2 m}{200t}}.$$

In combination with (39) and (40), this implies that for $m \geq 200t(2d+1)/\epsilon^2$

$$\mathbb{P}\left[ A \right] \leq \frac{\delta}{2} + 8 \left( 16e^2 \right)^d e^{-\frac{\epsilon^2 m}{200t}}.$$

By the main claim, we always have $m \geq 200t(2d+1)/\epsilon^2$ and hence we only need to show that

$$\mathbb{P}\left[ A \right] \leq \frac{\delta}{2} + 8 \left( 16e^2 \right)^d e^{-\frac{\epsilon^2 m}{200t}} \leq \delta.$$

This is equivalent to

$$8 \left( 16e^2 \right)^d e^{-\frac{\epsilon^2 m}{200t}} \leq \delta/2$$

and

$$\log 16 + d(\log 16 + 2) - \frac{\epsilon^2 m}{200t} \leq \ln \delta.$$

This is the case if we have

$$\log 16 + d(\log 16 + 2) + \log \frac{1}{\delta} \leq \frac{\epsilon^2 m}{200t}$$

or equivalently

$$m \geq \frac{200t}{\epsilon^2} \left( \log 16 + d(\log 16 + 2) + \log \frac{1}{\delta} \right).$$

The main claim thus holds since

$$m \geq \frac{200t}{\epsilon^2} \left( 3 + 5d + \log \frac{1}{\delta} \right)$$

which concludes the proof. □

# C. Proof of Lemma 1

*Proof.* We first show (9) with the same notation as in the proof of Lemma 4. For any $f \in \mathcal{F}$, $\vec{x} \in \mathcal{X}^m$ and $\vec{r} \in \mathbb{R}^m$, we denote the restriction $(f(x_1), \ldots, f(x_m))$ by $f_{\vec{x}}$ and set $\mathcal{F}_{\vec{x}} = \{f_{\vec{x}} \mid f \in \mathcal{F}\}$. For any vector $\vec{z} \in \mathbb{R}^m$, we define

$$\mathrm{sign}(\vec{z}) = (\mathrm{sign}(z_1), \ldots, \mathrm{sign}(z_m)).$$

The set of dichotomies induced by $\vec{r}$ on $F_{\vec{x}}$ is given by

$$\mathrm{sign}(\mathcal{F}_{\vec{x}} - \vec{r}) = \{\mathrm{sign}(f_{\vec{x}} - \vec{r}) \mid f \in \mathcal{F}\}.$$

Let $\bar{m}$ be equal to the pseudo-dimension of $\mathcal{F}$. This implies that there exist two vectors $\bar{x} \in \mathcal{X}^{\bar{m}}$ and $\bar{r} \in \mathbb{R}^{\bar{m}}$ that are shattered by $\mathcal{F}$, i.e.,

$$|\mathrm{sign}(\mathcal{F}_{\bar{x}} - \bar{r})| = 2^{\bar{m}}. \tag{41}$$

Consider any $x \in \vec{x}$, its corresponding $r \in \vec{r}$ and any $f_Q(\cdot) \in \mathcal{F}$. Defining $\sigma_Q^2 = \mathbb{E}_P\left[\mathrm{d}(x,Q)^2\right]$, we have that

$$\begin{aligned}
&\mathrm{sign}(f_Q(x) - r_x) \\
&= \mathrm{sign}\left(\frac{\mathrm{d}(x,Q)^2}{\frac{1}{2}\sigma^2 + \frac{1}{2}\mathbb{E}_P\left[\mathrm{d}(x,Q)^2\right]} - r\right) \\
&= \mathrm{sign}\left(\mathrm{d}(x,Q)^2 - \frac{r_x}{2}\left(\sigma^2 + \sigma_Q^2\right)\right) \\
&= \mathrm{sign}\left(\min_{q \in Q} \mathrm{d}(x,q)^2 - \frac{r}{2}\left(\sigma^2 + \sigma_Q^2\right)\right) \\
&= \mathrm{sign}\left(\min_{q \in Q}\left[x^T x - 2x^T q + q^T q\right] - \frac{r}{2}\left(\sigma^2 + \sigma_Q^2\right)\right) \\
&= \mathrm{sign}\left(\min_{\tilde{q} \in \tilde{Q}(Q)} \langle \tilde{x}(x,r), \tilde{q}\rangle\right),
\end{aligned} \tag{42}$$

where we have used the mappings

$$\tilde{x}(x,r) = \begin{pmatrix} -2x \\ -r/2 \\ 1 \\ x^T x \end{pmatrix}$$

and

$$\tilde{Q}(Q) = \left\{\begin{pmatrix} q \\ \sigma^2 + \sigma_Q^2 \\ q^T q \\ 1 \end{pmatrix} \,\Big|\, q \in Q\right\}.$$

Consider the vector $\tilde{x} = (\tilde{x}(\bar{x}_1, \bar{r}_1), \ldots, \tilde{x}(\bar{x}_{\bar{m}}, \bar{r}_{\bar{m}}))$ and the function family

$$\mathcal{G} = \left\{\min_{\tilde{q} \in \tilde{Q}(Q)} \langle \cdot, \tilde{q}\rangle \,\Big|\, Q \in \mathbb{R}^{d \times k}\right\}.$$

By (42), both $\mathcal{F}$ and $\mathcal{G}$ induce the same dichotomies on $(\bar{x}, \bar{r})$ and $(\tilde{x}, \vec{0})$ respectively and thus

$$|\mathrm{sign}(\mathcal{F}_{\bar{x}} - \bar{r})| = |\mathrm{sign}(\mathcal{G}_{\tilde{x}})|. \tag{43}$$

We define the function family

$$\mathcal{H} = \left\{\min_{\tilde{q} \in Q} \langle \cdot, \tilde{q}\rangle \,\Big|\, Q \in \mathbb{R}^{(d+3) \times k}\right\}$$

and note that by construction $\mathcal{G} \subseteq \mathcal{H}$. This implies

$$|\mathrm{sign}(\mathcal{G}_{\tilde{x}})| \leq |\mathrm{sign}(\mathcal{H}_{\tilde{x}})|. \tag{44}$$

Consider the function family

$$\mathcal{I} = \left\{\langle \cdot, \tilde{q}\rangle \mid \tilde{q} \in \mathbb{R}^{d+3}\right\}.$$

For any $Q \in \mathbb{R}^{(d+3) \times k}$, it holds that

$$\left\{x \in \tilde{x} \mid \min_{\tilde{q} \in Q} \langle x, \tilde{q}\rangle \leq 0\right\} = \bigcup_{\tilde{q} \in Q} \{x \in \tilde{x} \mid \langle x, \tilde{q}\rangle \leq 0\}.$$

Since $|Q| = k$, this implies that there exists an injective mapping from $\mathcal{H}$ to the $k$-fold Cartesian product of $\mathcal{I}$ that generates the same dichotomies. In turn, this implies

$$|\mathrm{sign}(\mathcal{H}_{\tilde{x}})| \leq |\mathrm{sign}(\mathcal{I}_{\tilde{x}})|^k. \tag{45}$$

The dichotomies induced by $\mathcal{I}$ are generated by halfspaces in $\mathbb{R}^{d+3}$. The Vapnik-Chervonenkis dimension of halfspaces in $\mathbb{R}^{d+3}$ is bounded by $d+4$ (Har-Peled, 2011) and thus $\mathrm{Pdim}(\mathcal{I}) \leq d+4$. Together with Sauer's Lemma (Sauer, 1972; Vapnik & Chervonenkis, 1971), this implies

$$|\mathrm{sign}(\mathcal{I}_{\tilde{x}})| \leq \left(\frac{e\bar{m}}{d+4}\right)^{d+4}. \tag{46}$$

Combining (41), (43), (44), (45) and (46) yields

$$2^{\bar{m}} \leq \left(\frac{e\bar{m}}{d+4}\right)^{(d+4)k}.$$

This implies that

$$\frac{\bar{m}}{d+4} \leq \frac{k}{\log 2}\left(1 + \log\frac{\bar{m}}{d+4}\right) \leq \frac{2k}{\log 2}\log\frac{\bar{m}}{d+3}$$

Since $\frac{\bar{m}}{d+4} > 0$ and $\frac{2k}{\log 2} > 0$, Lemma 2 implies that

$$\frac{\bar{m}}{d+4} \leq \frac{4k}{\log 2}\log\frac{4k}{\log 2}$$

Since $\frac{4}{\log 2} \approx 5.77 < 6$, this proves the claim in (9), i.e.,

$$\mathrm{Pdim}(\mathcal{F}) = \bar{m} \leq 6k(d+4)\log 6k.$$

Next, we prove (10). For any $x \in \mathbb{R}^d$ and $Q \in \mathbb{R}^{d \times k}$, we have by the triangle inequality

$$\mathrm{d}(x, Q)^2 \leq (\mathrm{d}(x, \mu) + \mathrm{d}(\mu, Q))^2$$
$$= \mathrm{d}(x, \mu)^2 + \mathrm{d}(\mu, Q)^2 + 2\,\mathrm{d}(x, \mu)\,\mathrm{d}(\mu, Q)$$

For any $0 \leq a \leq b$, it holds that

$$2ab = ab + a(b-a) + a^2 \leq ab + b(b-a) + a^2 = b^2 + a^2.$$

Since either $0 \leq \mathrm{d}(x,\mu) \leq \mathrm{d}(\mu, Q)$ or $0 \leq \mathrm{d}(\mu, Q) < \mathrm{d}(x, \mu)$, we thus have for any $x \in \mathbb{R}^d$ and $Q \in \mathbb{R}^{d \times k}$

$$\mathrm{d}(x, Q)^2 \leq 2\,\mathrm{d}(x, \mu)^2 + 2\,\mathrm{d}(\mu, Q)^2. \qquad (47)$$

By the same argument it also holds that for any $x \in \mathbb{R}^d$ and $Q \in \mathbb{R}^{d \times k}$

$$\mathrm{d}(\mu, Q)^2 \leq 2\,\mathrm{d}(x, \mu)^2 + 2\,\mathrm{d}(x, Q)^2.$$

By taking the expectation with regards to $P$ and noting that $\sigma^2 = \mathbb{E}_P\left[\mathrm{d}(x,\mu)^2\right] < \infty$, we obtain for any $Q \in \mathbb{R}^{d \times k}$

$$\mathrm{d}(\mu, Q)^2 \leq 2\sigma^2 + 2\mathbb{E}_P\left[\mathrm{d}(x, Q)^2\right]. \qquad (48)$$

Combining (47) and (48) implies that for any $x \in \mathbb{R}^d$ and $Q \in \mathbb{R}^{d \times k}$

$$\mathrm{d}(x, Q)^2 \leq 2\,\mathrm{d}(x, \mu)^2 + 4\sigma^2 + 4\mathbb{E}_P\left[\mathrm{d}(x, Q)^2\right]$$
$$\leq \left(4 + \frac{2\,\mathrm{d}(x, \mu)^2}{\sigma^2}\right)\sigma^2 + 4\mathbb{E}_P\left[\mathrm{d}(x, Q)^2\right]$$
$$\leq \left(4 + \frac{2\,\mathrm{d}(x, \mu)^2}{\sigma^2}\right)\left(\sigma^2 + \mathbb{E}_P\left[\mathrm{d}(x, Q)^2\right]\right)$$
$$\leq \left(8 + \frac{4\,\mathrm{d}(x, \mu)^2}{\sigma^2}\right)\frac{1}{2}\left(\sigma^2 + \mathbb{E}_P\left[\mathrm{d}(x, Q)^2\right]\right).$$

By the definition of $f_Q(x)$, this proves (10). Finally we have

$$\mathbb{E}_P\left[s(x)^2\right] = \mathbb{E}_P\left[\left(\frac{4\,\mathrm{d}(x, \mu)^2}{\sigma^2} + 8\right)^2\right]$$
$$= \mathbb{E}_P\left[\left(\frac{16\,\mathrm{d}(x,\mu)^2}{\sigma^2} + \frac{64\,\mathrm{d}(x,\mu)^2}{\sigma^2} + 64\right)\right]$$
$$= 128 + 16\frac{\mathbb{E}_P\left[\mathrm{d}(x,\mu)^4\right]}{\sigma^2}$$
$$= 128 + 16\hat{M}_4.$$

which shows (11) and concludes the proof. $\square$